\documentclass[sigconf,preprint]{acmart}

\setcopyright{none}
\renewcommand\footnotetextcopyrightpermission[1]{}
\acmConference[KDD '27]{Proceedings of the 33rd ACM SIGKDD Conference on Knowledge Discovery and Data Mining}{August 1--5, 2027}{San Jose, CA, USA}

\usepackage{booktabs}
\usepackage{multirow}
\usepackage{colortbl}
\usepackage{array}
\usepackage{tabularx}
\usepackage{graphicx}
\usepackage{subcaption}
\usepackage{xcolor}
\usepackage{amsmath}
\usepackage{enumitem}
\usepackage{balance}
\usepackage{multicol}
\usepackage{listings}
\usepackage{tikz}
\usepackage{url}
\usepackage{tcolorbox}
\usetikzlibrary{arrows.meta,positioning,shapes.geometric,fit,calc,backgrounds}

\tcbuselibrary{listings,raster}

\definecolor{codebg}{HTML}{F7F7F8}
\definecolor{codeframe}{HTML}{DADCE0}
\definecolor{promptbg}{HTML}{F3F0FF}
\definecolor{promptframe}{HTML}{7E57C2}
\definecolor{goodbg}{HTML}{E8F5E9}
\definecolor{goodframe}{HTML}{2E7D32}
\definecolor{badbg}{HTML}{FBE9E7}
\definecolor{badframe}{HTML}{C62828}
\definecolor{l5bg}{HTML}{E3F2FD}
\definecolor{l5frame}{HTML}{1565C0}
\definecolor{lvl1frame}{HTML}{388E3C}
\definecolor{lvl2frame}{HTML}{1976D2}
\definecolor{lvl3frame}{HTML}{F57C00}
\definecolor{lvl4frame}{HTML}{C62828}
\definecolor{lvl5frame}{HTML}{7B1FA2}

\tcbset{
  emrb/.style={
    boxrule=0.5pt, arc=3pt, left=5pt, right=5pt, top=3pt, bottom=3pt,
    fonttitle=\small\bfseries\sffamily, coltitle=white,
  },
}

\newtcolorbox{promptbox}[1][]{
  emrb, colback=promptbg, colframe=promptframe, colbacktitle=promptframe,
  title={#1}}

\newtcblisting{codeprompt}[1][]{
  emrb, colback=codebg, colframe=codeframe,
  colbacktitle=promptframe, coltitle=white,
  listing only, listing options={basicstyle=\ttfamily\scriptsize, breaklines=true, numbers=none},
  title={#1}}

\newtcolorbox{goodbox}[1][]{
  emrb, colback=goodbg, colframe=goodframe, colbacktitle=goodframe,
  title={#1}}

\newtcolorbox{badbox}[1][]{
  emrb, colback=badbg, colframe=badframe, colbacktitle=badframe,
  title={#1}}

\newtcolorbox{l5box}[1][]{
  emrb, colback=l5bg, colframe=l5frame, colbacktitle=l5frame,
  title={#1}}

\newtcolorbox{lvlbox}[2]{
  emrb, colback=#1!8!white, colframe=#1, colbacktitle=#1,
  title={#2}}

\setlist{nosep,leftmargin=*}

\begin{document}

\title{EMRB: A Multi-Level Benchmark for Evaluating LLM Reasoning\\over Raw Electromagnetic Signals}

\author{Mingxu Zhang}
\affiliation{\institution{The Hong Kong University of Science and Technology (Guangzhou)}\city{Guangzhou}\country{China}}
\email{mzhang630@connect.hkust-gz.edu.cn}

\author{Ying Sun}
\affiliation{\institution{The 63rd Research Institute, National University of Defense Technology}\city{Nanjing}\country{China}}
\email{sunyinggilly@gmail.com}

\author{Yuhan Li}
\affiliation{\institution{The Hong Kong University of Science and Technology (Guangzhou)}\city{Guangzhou}\country{China}}
\email{yuhanli530@gmail.com}

\author{Yang Ji}
\affiliation{\institution{The Hong Kong University of Science and Technology (Guangzhou)}\city{Guangzhou}\country{China}}
\email{yangji.me721@gmail.com}

\author{Dazhong Shen}
\affiliation{\institution{Nanjing University of Aeronautics and Astronautics}\city{Nanjing}\country{China}}
\email{shendazhong@nuaa.edu.cn}

\author{Ke Zhang}
\affiliation{\institution{The 63rd Research Institute, National University of Defense Technology}\city{Nanjing}\country{China}}
\email{zhangke24@nudt.edu.cn}

\author{Shan Huang}
\affiliation{\institution{The 63rd Research Institute, National University of Defense Technology}\city{Nanjing}\country{China}}
\email{huangshan12@nudt.edu.cn}

\begin{abstract}
Large language models (LLMs) are increasingly used as code agents for scientific and engineering analysis, but their ability to analyze raw physical-layer measurements remains untested.
We introduce \textbf{EMRB} (\textbf{E}lectro\textbf{m}agnetic \textbf{R}easoning \textbf{B}enchmark), which evaluates whether LLMs can analyze raw I/Q data by writing and running code.
EMRB contains 200 problems across five difficulty levels and 27 question types, from signal detection to OFDM design, generated from 11 signal types with verified ground truth.
Unlike benchmarks built on preprocessed features or structured tables, EMRB provides only the raw capture; the quantities each question refers to must first be discovered through code.
We evaluate 14 LLMs spanning proprietary, open-weight, and reasoning-oriented families.
Scores range from 24.1\% to 78.9\%, with the mean dropping from 84.9\% on basic measurement to 21.2\% on system design.
We also propose \textbf{ReconPilot}, a structured method that separates signal reconnaissance, targeted analysis, and self-verification.
Across three backbones, ReconPilot raises the overall score by 3.8 to 17.6 points and improves 13 of 15 backbone-level combinations tested.
All data and code are publicly released in \href{https://github.com/mingxuZhang2/EMRB}{\textcolor{blue}{our GitHub repository}}.
\end{abstract}


\maketitle

\section{Introduction}\label{sec:intro}

Electromagnetic signal analysis underpins wireless communications, spectrum management, radar, and electronic warfare.
Engineers routinely extract measurements from noisy I/Q captures to monitor spectrum, diagnose interference, and design systems, tasks that demand both signal-processing expertise and multi-step numerical reasoning.

LLMs offer a natural interface for such analysis: an LLM agent can inspect an unfamiliar capture, write Python code to compute the relevant measurements, and revise its analysis over multiple turns.
However, no systematic benchmark evaluates whether general-purpose LLMs can actually analyze raw electromagnetic signals this way.
Existing EM benchmarks~\cite{oshea2016radioml,merlin2026} rely on task-specific encoders or classify short snippets, and none require the model to discover what signals are present, implement analysis code, and answer engineering questions from the raw data.

A raw I/Q capture exposes no schema.
In tabular benchmarks, variables are named by columns; in code generation, the target behavior is specified in the prompt.
The variables an engineering question refers to, such as emitters, carriers, bandwidths, and interference relationships, do not exist in the input until the model computes them, and every downstream conclusion inherits the errors of this discovery step.

We introduce \textbf{EMRB} with 200 problems across five difficulty levels and 27 question types, generated from 11 signal types with verified ground truth.
Each problem provides a raw I/Q file and requires the model to write and execute Python analysis code rather than respond from preprocessed features.
We evaluate 14 LLMs and find scores ranging from 24.1\% to 78.9\%.
Performance declines sharply as tasks require more connected measurements: isolated measurement and textbook calculation can already be delegated to frontier models, while multi-signal analysis and system design cannot.
The bottleneck is not domain knowledge but the ability to keep each computed variable attached to the correct entity as the analysis chain grows.
We further propose \textbf{ReconPilot}, a structured method that separates fixed signal reconnaissance, targeted analysis, and self-verification, raising the overall score by up to 17.6 points across three backbones.

Our contributions are threefold:
\begin{enumerate}
  \item We introduce EMRB, a 200-problem benchmark across five difficulty levels and 27 question types with raw I/Q files, verified ground truth, and a code-execution protocol, covering tasks from signal detection to OFDM design.
  All signals are programmatically generated with deterministic scoring, so every answer is objectively verifiable.

  \item We propose ReconPilot, a three-stage method that separates fixed signal reconnaissance, targeted LLM analysis, and self-verification.
  Across three backbones, ReconPilot raises the overall score by up to 17.6 points, improving 13 of 15 backbone-level combinations tested, showing that a fixed reconnaissance stage can substitute for capability the backbone lacks (\S\ref{sec:reconpilot_results}).

  \item We evaluate 14 LLMs across proprietary, open-weight, and reasoning-oriented families, finding scores from 24.1\% to 78.9\% and a sharp decline from basic measurement (84.9\%) to system design (21.2\%).
\end{enumerate}

\vspace{-2pt}
\section{Related Work}\label{sec:related}

\textbf{EM signal benchmarks.}
Existing EM and wireless datasets such as RadioML~\cite{oshea2016radioml}, WiSig~\cite{hanna2022wisig}, WASD~\cite{kim2024wasd}, DeepMIMO~\cite{alkhateeb2019deepmimo}, and DeepSense 6G~\cite{charan2022deepsense} evaluate fixed outputs (modulation labels, device IDs, beam indices) under supervised settings, without asking an LLM to inspect raw measurements, write code, and interpret results.
MERLIN's EM-Bench~\cite{merlin2026} moves closer to language-based evaluation but still relies on a specialized signal encoder rather than code-mediated analysis, and TeleCom-Bench~\cite{xiao2026telecombench} targets textual comprehension rather than raw signal data.
EMRB complements these datasets by requiring analysis workflows over raw I/Q measurements; Table~\ref{tab:benchmark_comparison} in Appendix~\ref{app:benchmark_table} summarizes the differences.

\textbf{Code-execution and data analysis benchmarks.}
Code generation benchmarks (HumanEval~\cite{chen2021humaneval}, MBPP~\cite{austin2021mbpp}, APPS~\cite{hendrycks2021apps}, DS-1000~\cite{lai2023ds1000}) and agent benchmarks (AgentBench~\cite{liu2024agentbench}, ToolBench~\cite{qin2023toolbench}, GAIA~\cite{mialon2024gaia}, MLAgentBench~\cite{huang2023mlagentbench}) evaluate code synthesis and multi-step tool use but are not grounded in raw physical measurements.
Data-analysis benchmarks such as DA-Bench~\cite{dabench2024} and TableBench~\cite{tablebench2024} are closer, but operate on structured tables with explicit schemas, whereas raw I/Q captures are complex-valued time series with no schema, so the model must construct its own variables through signal processing before answering.

\textbf{Deep learning for EM signal processing.}
Deep learning methods for EM signals, from modulation classifiers~\cite{oshea2016convolutional} to end-to-end physical layer systems~\cite{oshea2017introduction} and joint channel estimation~\cite{ye2018power} (see~\cite{zhang2019deep} for a survey), achieve high accuracy but each addresses a single task and requires retraining for new signal types.
More recently, LLMs have been coupled to EM systems through learned encoders~\cite{zhou2025llmtelecom}: RF-GPT~\cite{zou2026rfgpt} processes spectrograms via a vision-language model, RadioLLM~\cite{chen2025radiollm} reprograms a frozen LLM for cognitive radio, and PReD~\cite{han2026pred} trains a multimodal foundation model for electromagnetic perception.
EMRB takes a complementary perspective: it evaluates whether general-purpose LLMs can analyze raw I/Q files directly through code execution without domain-specific fine-tuning.

\section{EMRB Benchmark Design}\label{sec:benchmark}

\begin{figure*}[t]
\centering
\includegraphics[width=0.96\textwidth]{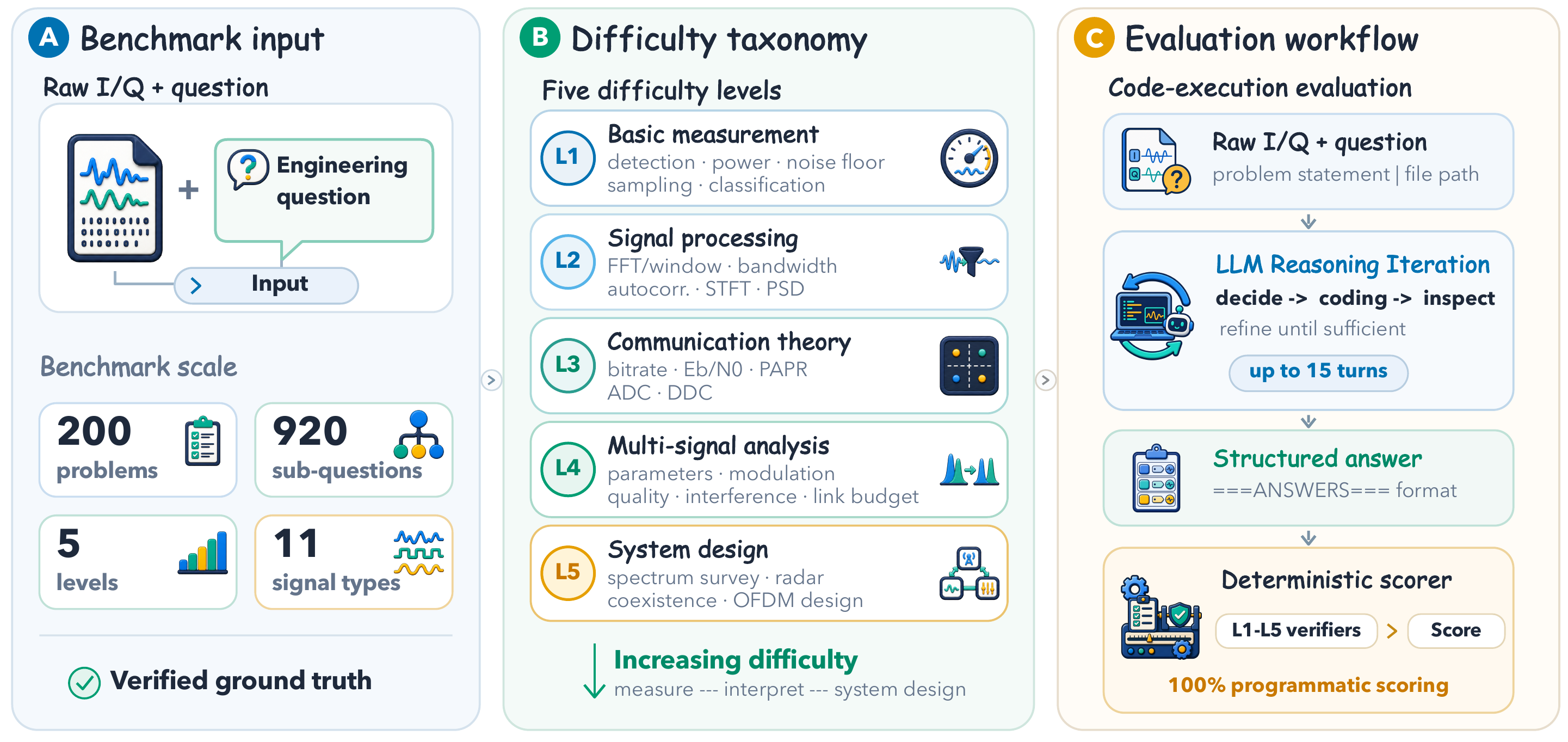}
\caption{Overview of EMRB. \textbf{(A)}~Benchmark input: raw I/Q capture paired with an engineering question (200 problems, 11 signal types, verified ground truth). \textbf{(B)}~Five difficulty levels from basic measurement to system design. \textbf{(C)}~Code-execution evaluation: the model writes and runs Python over up to 15 turns, scored by deterministic verifiers.}
\label{fig:overview}
\end{figure*}

\subsection{Design Principles}\label{sec:principles}

Figure~\ref{fig:overview} gives an overview of the benchmark.
EMRB's design follows three principles.
First, the evaluation must preserve an authentic analysis workflow: load raw I/Q data, choose signal-processing methods, implement them in code, and use numerical results to answer the question.
Second, the benchmark must expose different stages of competence from basic measurements to system-level reasoning across a broad range of EM analysis problems.
Third, signals are generated programmatically with fixed random seeds and known ground truth, so every problem is reproducible and every answer is objectively verifiable.

\subsection{Code-Execution Evaluation Protocol}\label{sec:code_exec}

EMRB provides each signal as a raw numerical data file rather than serializing I/Q samples into the prompt.
Raw electromagnetic measurements are not naturally represented as text: an I/Q capture is a complex-valued array whose analysis requires numerical procedures such as FFT, PSD estimation, autocorrelation, and peak detection, which must be executed on arrays with controlled precision.
EMRB therefore treats the signal file as data to be loaded and analyzed through code.
The prompt contains the question and the file path; the model writes Python analysis code, observes numerical outputs, and refines its answers over multiple turns.
The concrete tool-calling loop, execution limits, and scoring procedure are described in Section~\ref{sec:eval}.

\subsection{Difficulty Taxonomy}\label{sec:taxonomy}

Raw EM analysis is not a single capability: a model may estimate the power of one isolated signal but fail when it must separate overlapping emitters, choose a bandwidth definition, or combine several measurements into a system-level decision.
EMRB organizes its 200 problems into five difficulty levels (Table~\ref{tab:levels} in Appendix~\ref{app:examples}), each containing 40 problems and targeting a progression from isolated measurements to integrated engineering judgment.
Across these levels, the benchmark defines 27 non-overlapping question types so that each level introduces new analytical requirements rather than repeating the same task at higher parameter complexity.
Representative problems for each level are provided in Appendix~\ref{app:examples}, and Table~\ref{tab:levels} lists the question types per level.

\textbf{L1 (Basic Measurement)} covers the first quantities an analyst reads off a raw I/Q capture once a coarse picture of the band is already in hand: at what frequency each source sits, how strong it is, where the noise floor lies, and what modulation is being used.
The prompt supplies a preliminary scan summary with source count and approximate frequencies, so L1 measures calibration precision against a known scene rather than detection from scratch; L2 and L3 open with the same summary, which is withdrawn from L4 onward (Appendix~\ref{app:full_problem}).
Each task can typically be solved with one analysis step, such as computing a PSD or locating spectral peaks.
L1 establishes whether a model can turn a rough prior into numerically reliable measurements before any deeper analysis begins.

\textbf{L2 (Signal Processing Methodology)} asks the model to choose the right measurement procedure, not just execute one.
Its five question types cover FFT/window selection, bandwidth measurement, autocorrelation, STFT analysis, and signal energy/power relationships.
The difficulty lies in the fact that the same signal can require different analyses depending on the question: bandwidth alone may refer to a 3-dB width, a null-to-null width, or an occupied-power width, each needing a different implementation.

\textbf{L3 (Communication Theory)} moves from measurement to interpretation.
The model must connect measured properties to communication-system quantities: bitrate, $E_b/N_0$, PAPR, ADC quantization, and digital down-conversion.
L1 and L2 characterize what is in the signal; L3 asks what it means for the communication link.

\textbf{L4 (Multi-Signal Analysis)} places the model in a crowded spectrum with three to five emitters.
Each problem draws five sub-questions from nine question types spanning symbol-rate estimation, analog and radar parameter extraction, burst analysis, spectral-gap link budgeting, interference assessment, and Shannon capacity.
Before answering any of them, the model must first build a signal inventory: how many emitters are present, where they are, and how to separate them.
A missed or merged emitter propagates errors to every downstream measurement.

\textbf{L5 (System Design)} requires the model to turn measurements into engineering decisions.
Its three question types cover spectrum survey, radar coexistence, and OFDM design.
Unlike L1 to L4, L5 demands sequential reasoning across the full analysis chain: the signal inventory feeds interference assessment, which feeds system design.
An error at any stage invalidates the final design.

\subsection{Signal Library and Data Quality}\label{sec:signal_lib}

EMRB uses a parameterized signal library covering 11 signal types: seven digital modulations (BPSK, QPSK, 8PSK, 16QAM, 64QAM, 2FSK, 4FSK), two analog (FM, AM-DSB), and two wideband (OFDM, LFM chirp).
For each signal, carrier frequency, bandwidth, power, and SNR are sampled from level-specific ranges and the waveform is synthesized with calibrated AWGN; for L4 and L5, several emitters share one capture, producing crowded spectra whose true inventory remains exactly known.
Each problem is uniquely determined by an (archetype, seed) pair: eight scenario archetypes per level fix the signal composition and question selection, and varying the seed produces a structurally identical but numerically distinct instance, so the released 200 problems (5 seeds $\times$ 8 archetypes $\times$ 5 levels) can be expanded to fresh, valid problems from unused seeds without manual authoring, each carrying automatically verified ground truth.
The dataset totals 920 sub-questions; Appendix~\ref{app:datasheet} gives per-type parameters, coverage limits, and the full generation pipeline.

Because every ground-truth answer comes from these generation parameters, EMRB verifies that each scored quantity is independently recoverable from the released waveform: reference signal-processing scripts re-measure every answer from the saved file and agree to within 1\% on all 200 problems.
A separate implementation-tolerance check confirms that each waveform realizes its target parameters (power $\pm$0.5\,dB, SNR $\pm$1\,dB, bandwidth $\pm$5\%), and a coverage check verifies the intended question-type and signal-type distribution.
Appendix~\ref{app:datasheet} gives the complete validation procedures.

\section{Evaluation Protocol}\label{sec:eval}

\subsection{Code-Execution Loop}\label{sec:agent_loop}

EMRB evaluates models in an interactive code-execution setting that mirrors how raw I/Q measurements are analyzed in practice: no signal is rendered as text, and every reported quantity must be computed from the binary file.
Each model receives the question and the raw I/Q file path with a sandboxed Python tool, and may take up to 15 interaction turns, rather than 15 code executions, before submitting a structured final answer.
The budget counts turns rather than code calls because real signal analysis often requires intermediate reasoning, such as inspecting a PSD plot and deciding which sub-band to zoom into, that does not involve code.
This design makes the benchmark test a complete analysis workflow rather than isolated subtasks: the model must decide what to compute, write executable signal-processing code, interpret numerical outputs, and judge when its measurements are sufficient.
Forcing the model to commit to a structured answer format also prevents hedging with qualitative language when a precise number is required.
Implementation details, including system prompt, sandbox configuration, answer format specification, and forced-generation fallback, are provided in Appendix~\ref{app:prompts}; Appendix~\ref{app:full_problem} (Figure~\ref{fig:full_prompt}) shows a complete problem prompt as delivered to a model.

\subsection{Deterministic Scoring System}\label{sec:scoring}

A benchmark built on raw physical measurements requires scoring that is both precise and reproducible.
EMRB achieves this through level-specific deterministic verifiers in which no LLM judge contributes to the reported scores.
L1 through L4 each contain five sub-questions worth 20 points; L5 contains three integrated sub-questions worth 34, 33, and 33 points.
Every scored requirement is checked against measured ground truth, an explicit feasibility constraint, or, for the prompts that ask for a recommendation or an explanation, a negation-aware assertion check on the committed conclusion.
Criteria that credit the content of a stated rationale rather than its conclusion amount to 11 of L2's 100 rubric points (2.2\% of the benchmark total), and do not appear at any other level; Appendix~\ref{app:rubrics} specifies their mechanism, weights, and limits.

Two design choices are critical to scoring validity.
First, the verifier performs \emph{entity-bound} evaluation: each extracted value is matched to its sub-question, physical unit, and signal identity before a quantity-specific tolerance is applied, so an unrelated number elsewhere in the response cannot receive credit.
Second, integrated L4 and L5 answers such as a spectral extraction passband, a channel packing plan, or an OFDM design are scored against functional acceptance rules rather than a single reference string, permitting multiple valid designs while preserving objectivity.
L5 additionally employs prerequisite-gated scoring: downstream sub-parts receive credit only when their upstream prerequisites are met (e.g., overlap analysis is gated on correct signal-pair identification), because a confident design applied to the wrong target is worse than no answer.
Verifier versions and task provenance fingerprints are stored with every result, so stale scores are automatically rejected when any question, waveform, or scoring rule changes.
The complete scoring rubrics are provided in Appendix~\ref{app:rubrics}.

\section{ReconPilot: Structured Signal Analysis}\label{sec:reconpilot}

\begin{figure}[t]
\centering
\includegraphics[width=\columnwidth]{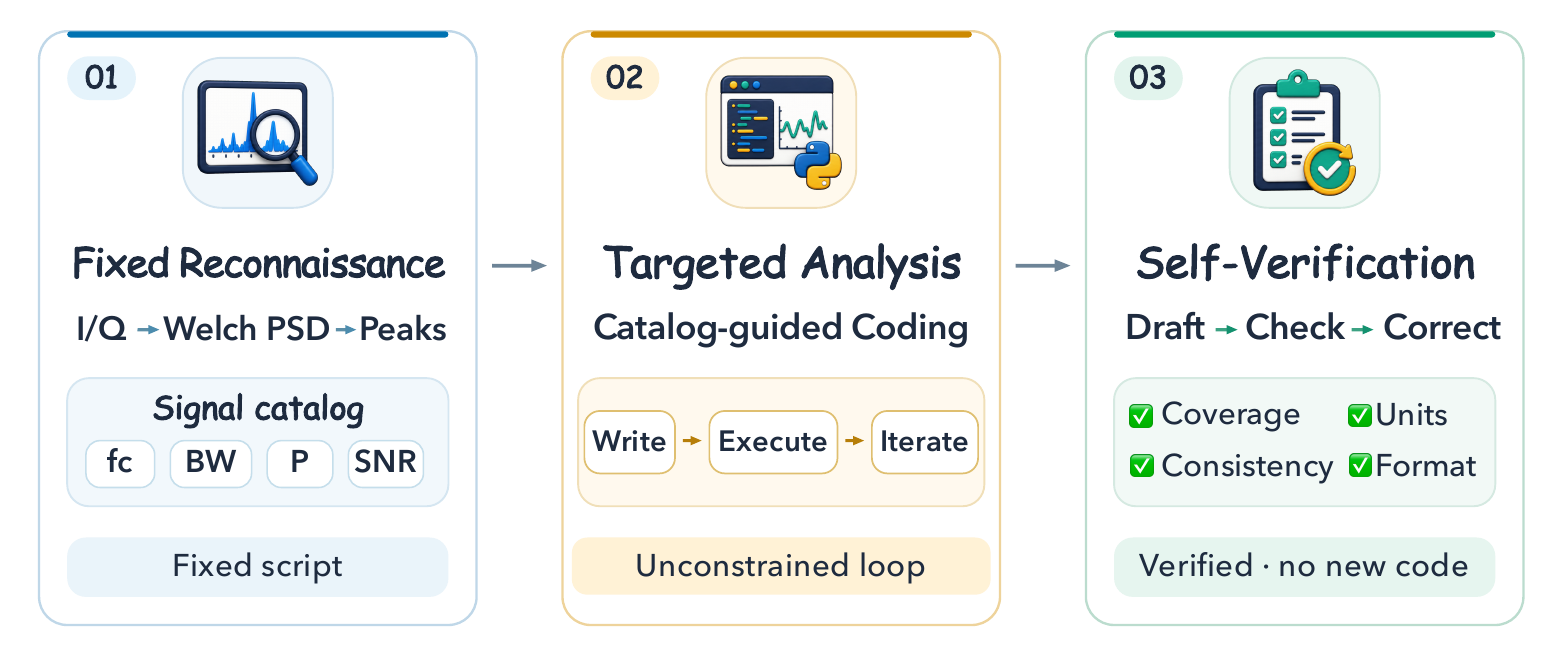}
\caption{ReconPilot three-stage architecture. Stage~1 turns raw I/Q into a bounded but deliberately unresolved spectral region map; Stage~2 guides targeted LLM analysis; Stage~3 verifies consistency before submission.}
\label{fig:pipeline}
\end{figure}

When an LLM analyzes a raw I/Q capture, it must simultaneously discover the signal scene and solve the task, and each half invites a distinct failure: before analysis begins, models often launch into undirected exploration without first understanding the scene; after it ends, they rarely check whether their own conclusions are internally consistent, e.g., a reported bitrate that contradicts the identified modulation, or a sub-question silently left unanswered.
ReconPilot targets these two failures by wrapping LLM-guided targeted analysis with a fixed stage on each side (Figure~\ref{fig:pipeline}): fixed reconnaissance supplies a common starting point, and self-verification catches inconsistencies before submission.
The method can be applied on top of any code-capable LLM.

\subsection{Three-Stage Pipeline Design}\label{sec:pipeline_design}

\textbf{Stage 1: Fixed Reconnaissance.}
Without any prior context, a model's first turns are typically spent on broad spectral sweeps that duplicate work across problems; providing a consistent starting picture eliminates this variance.
A deterministic script, shared across all problems and backbones, estimates the noise floor via Welch PSD, thresholds at 6\,dB above the floor to detect occupied regions, and reports each region's frequency limits, width, integrated power, and peak level.
Per-region STFT statistics flag time-varying, broadband, or frequency-drifting behavior.
Critically, the script never counts sources, names modulations, or fits parameters: a region may contain one source, several overlapping emitters, or a swept signal, and the region count is explicitly not the source count.
A guessed inventory would anchor the model to a wrong decomposition, so Stage~1 bounds the search space without resolving it (Appendix~\ref{app:recon}).

\textbf{Stage 2: Reconnaissance-Guided Analysis.}
The model receives the region map from Stage~1 as additional context alongside the question and the raw I/Q file, and analyzes them with the same unconstrained code-execution loop used in the free-form baseline.
By skipping the repetitive initial spectral exploration that free-form runs spend their early turns on, the model can devote its full turn budget to task-specific measurement and reasoning.
The map localizes where to look but answers nothing by itself: the LLM must still determine source counts, select the appropriate measurement procedure, and carry out task-specific calculations.

\textbf{Stage 3: Self-Verification.}
During multi-turn code execution, models are prone to silent internal inconsistencies: a reported bitrate may contradict the identified modulation order, or a sub-question the model believed it had answered may in fact be missing.
To address this, Stage~3 requires the model to review its own answers before submission without executing new code, checking sub-question coverage, unit consistency, physical plausibility, and answer format compliance, converting the completed analysis into a verified final answer.


\begin{figure*}[!t]
  \centering
  \includegraphics[width=0.94\textwidth]{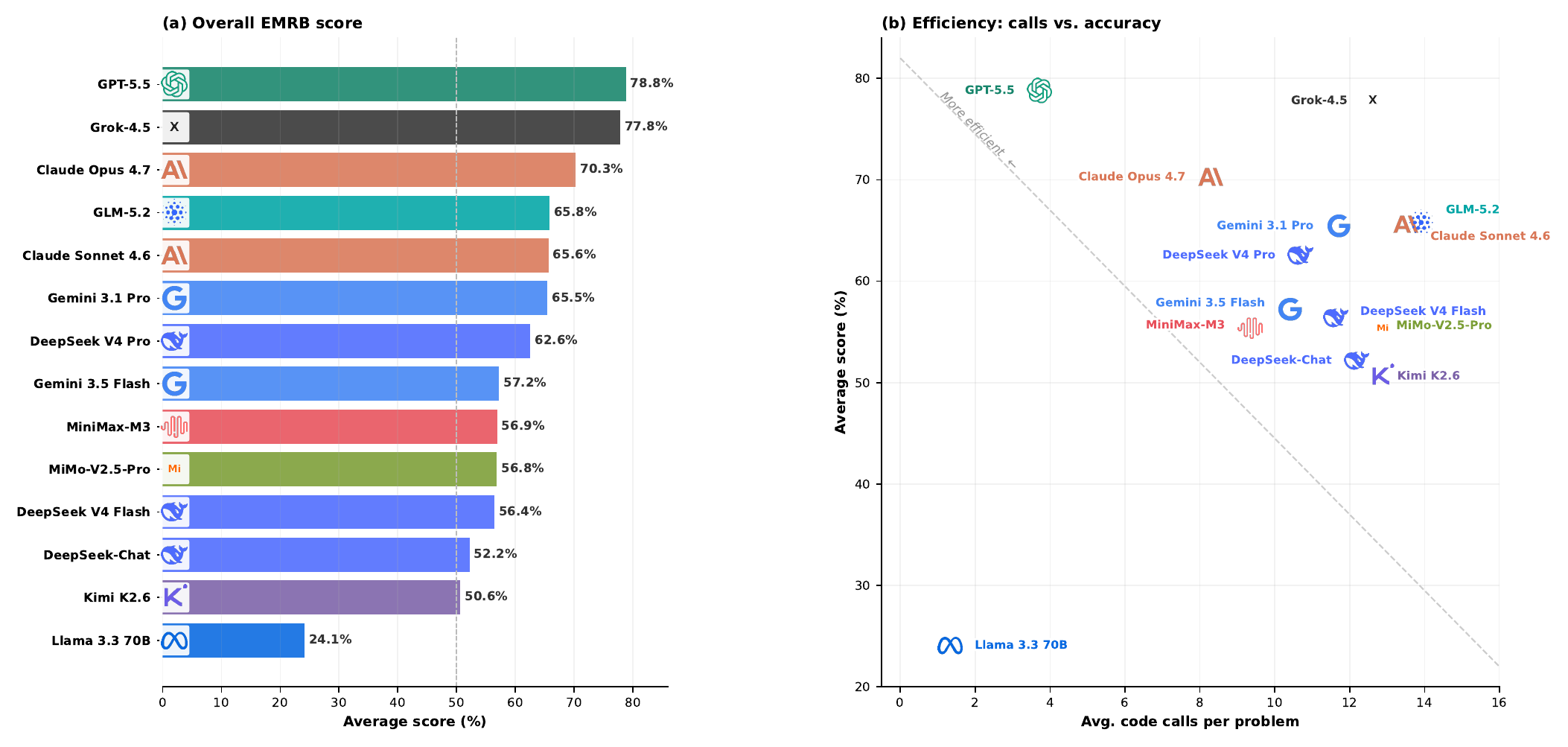}
  \caption{Overall EMRB results across 14 LLMs. (a)~Average scores over 200 problems. (b)~Code calls per problem versus accuracy. Colors and icons are shared between panels.}
  \label{fig:main_results}
\end{figure*}

\section{Experiments}\label{sec:experiments}

We organize the evaluation around three research questions.
\textbf{RQ1:} How much of EMRB can current LLMs solve, and is the remaining headroom achievable?
\textbf{RQ2:} Do EMRB's five levels test separable cognitive demands, and where do models systematically fail?
\textbf{RQ3:} Does staged reconnaissance improve raw-signal analysis, and how does each stage contribute?

\subsection{Experimental Setup}\label{sec:setup}

\textbf{Models evaluated.}
We evaluate 14 LLMs spanning proprietary, open weight, reasoning oriented, and lightweight model families: GPT-5.5~\cite{openai2026gpt55}, Grok-4.5~\cite{xai2026grok45}, Claude Opus 4.7~\cite{anthropic2026opus47} and Claude Sonnet 4.6~\cite{anthropic2026sonnet46}, Gemini 3.1 Pro~\cite{google2026gemini31} and Gemini 3.5 Flash~\cite{google2026gemini35flash}, GLM-5.2~\cite{zeng2026glm5}, DeepSeek V4 Pro, DeepSeek V4 Flash, and DeepSeek-Chat~\cite{deepseek2026v4}, MiniMax-M3~\cite{lai2026minimaxm3}, MiMo-V2.5-Pro~\cite{xiaomi2026mimo25,xiaomi2026mimo}, Kimi-K2.6~\cite{kimiteam2026kimik25}, and Llama 3.3 70B~\cite{grattafiori2024llama3}.
This pool covers a broad range of current code capable systems rather than only a small set of frontier models.
Every model receives the same problem inputs, raw file access, Python tool interface, 15 turn budget, and deterministic scoring procedure described in Section~\ref{sec:agent_loop}.
We additionally evaluate ReconPilot with DeepSeek V4 Flash, Gemini 3.5 Flash, and GPT-5.5 as backbones to compare unconstrained tool use with an explicitly staged analysis process under a fixed backbone.

\textbf{Configuration.}
All models are evaluated at temperature 0 with provider default reasoning settings.
We use the same interaction budget for all free form runs: each model receives the problem statement, the raw I/Q file path, and up to 15 turns before submitting a final answer.
ReconPilot follows the three stage procedure described in Section~\ref{sec:reconpilot}; its deterministic reconnaissance stage is shared across problems, while targeted analysis uses the same backbone model as the corresponding free form run before self-verification.
All reported scores are computed on the 200 EMRB problems using the deterministic scoring protocol in Section~\ref{sec:scoring}.

\begin{figure*}[!t]
  \centering
  \includegraphics[width=\textwidth]{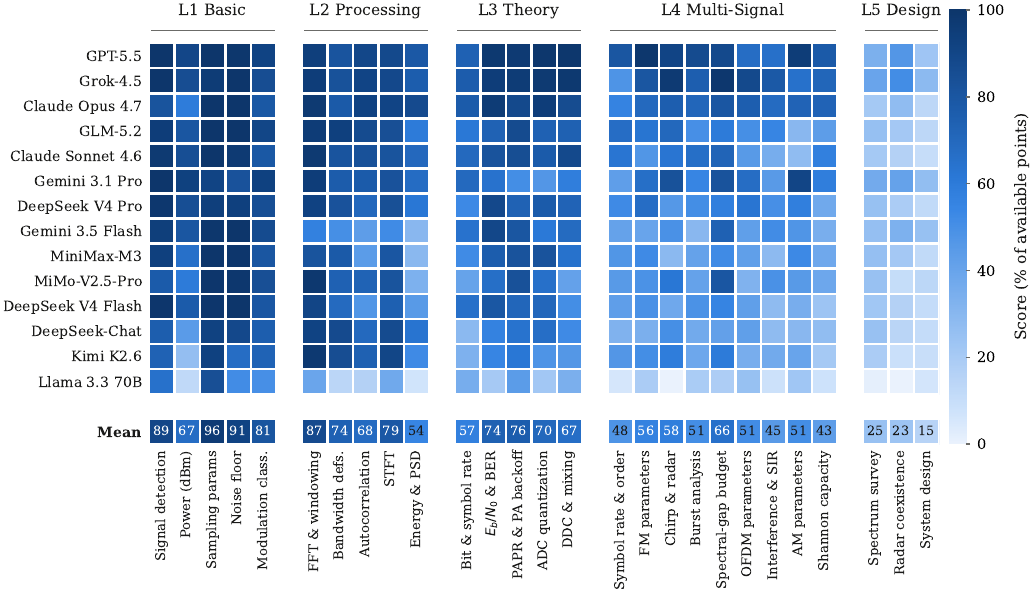}
  \caption{EMRB scores for all 27 question types across 14 models, expressed as a percentage of the points available for that type. Models are ordered by overall score; the bottom row gives the mean over models for each type.}
  \label{fig:question_types}
\end{figure*}

\subsection{Overall Performance}\label{sec:results}

To address RQ1, Figure~\ref{fig:main_results}(a) summarizes overall EMRB performance across the 14 evaluated models.
Complete per-level scores for all models are listed in Appendix~\ref{app:full_results}.
Scores range from 24.1\% to 78.9\%, a spread of 54.8 percentage points.
GPT-5.5 and Grok-4.5 form the leading group at 78.9\% and 77.8\%, followed by Claude Opus 4.7 at 70.3\%.
GLM-5.2, Claude Sonnet 4.6, Gemini 3.1 Pro, and DeepSeek V4 Pro score between 62.6\% and 65.8\%, while six further models occupy the 50.6\% to 57.2\% range.
Llama 3.3 70B scores 24.1\%, showing that access to the same Python tools does not eliminate large differences in the ability to organize raw signal analysis.
No model reaches 80\%, so the benchmark remains unsaturated even for the strongest current systems.

Figure~\ref{fig:main_results}(b) adds an efficiency dimension by plotting overall accuracy against average code calls per problem.
Notably, higher performance does not simply come from executing more code.
GPT-5.5 reaches 78.9\% with 3.7 code calls per problem, while Grok-4.5 reaches a similar 77.8\% with 12.6 calls, more than three times as many tool interactions for nearly equal accuracy.
EMRB does not reward brute-force code iteration; it tests whether a model can identify the right signal-processing method for the question, implement it correctly in few attempts, and interpret the numerical output to reach a sound and well-justified engineering conclusion.

Within model families, members share training corpora, tool-calling, and alignment pipelines.
The consistent gaps EMRB reveals across family members, 10.4 points across three DeepSeek variants, 8.3 between two Gemini models, and 4.7 between two Claude models, therefore isolate reasoning depth as the bottleneck.
The signal-processing knowledge needed for EMRB is present even in each family's lighter variants, but assembling that knowledge into a correct multi-step analysis plan requires the sustained inference chains that only flagship-scale architectures maintain.

Figure~\ref{fig:question_types} disaggregates the overall scores to all 27 question types, revealing structure that level averages hide.
All three L5 types score below 25 points averaged over models, confirming that the deficit is spread across the entire level rather than concentrated in one sub-question.
At L4, Shannon capacity is the weakest type at 43 despite being the most standard textbook formula in the level, indicating that the bottleneck lies in extracting the correct bandwidth and noise terms from a multi-emitter scene rather than in the calculation itself.
At L1, absolute power in dBm is the only weak type (67 versus 81--96 for the other four), and this weakness persists upward: the L2 energy and PSD type is also the weakest of its level at 54, suggesting that calibrated power measurement is the first capability to break and remains so across difficulty levels.

\subsection{Headroom Validity}\label{sec:headroom}

EMRB reports substantial headroom at every level and almost total headroom at L5, so before interpreting those numbers we must show that the missing points are reachable.
We have no human expert baseline (Section~\ref{sec:limitations}), and we therefore establish reachability with three converging measurements, summarized in Table~\ref{tab:headroom}.
We first replay the ground truth through the scorer itself.
For each problem we render the reference values in the answer format the model is asked to produce, then score that response with the same verifiers used for every reported result.
This yields 100.0 at all five levels, with 40 of 40 problems scored perfectly at each.
The rubric is therefore satisfiable everywhere, and the low L5 scores are not produced by rules that no response can satisfy.

We then take the union over models.
For every criterion on every problem we retain the best result achieved by any of the 14 models, which measures what the current field as a whole can do rather than what any single system does.
At L1 through L4 this union covers 98.8\% to 99.8\% of the available points, and all 143 criteria at those levels are solved perfectly by at least one model on at least one problem.
Even on these four levels the best single-model score ranges from 84.8\% to 95.8\%, well below the union, so the remaining gap is a difference in capability across models rather than a product of strict scoring.
At L5 the same union reaches only 53.2\%, and six criteria are never scored perfectly across 560 attempts, which we audit in Appendix~\ref{app:scoring_reliability}.

\begin{table}[t]
\centering
\caption{EMRB headroom is reachable. Oracle replay: ground truth scored by the verifier. Union of 14: best per-criterion result across models. Reference solver: classical DSP (L5 only).}
\label{tab:headroom}
\small
\begin{tabular}{lrrrr}
\toprule
\textbf{Level} & \textbf{Oracle} & \textbf{Union} & \textbf{Best} & \textbf{Reference} \\
               & \textbf{replay} & \textbf{of 14}  & \textbf{model} & \textbf{solver} \\
\midrule
L1 & 100.0 & 99.8 & 95.8 & -- \\
L2 & 100.0 & 99.4 & 88.7 & -- \\
L3 & 100.0 & 99.6 & 93.2 & -- \\
L4 & 100.0 & 98.8 & 84.8 & -- \\
L5 & 100.0 & 53.2 & 39.8 & \textbf{75.7} \\
\bottomrule
\end{tabular}
\end{table}

Because the union leaves half of L5 unaccounted for, we additionally wrote a reference solver for that level: roughly 1000 lines of classical DSP that read only the released waveform and the question text, with no access to generation parameters and no per-problem special casing.
It scores 75.7 out of 100, close to twice the best model score of 39.8, while declining by construction to answer 18 points worth of sub-questions; on the 82 points it does attempt it earns 92.3\%.
Two of the six criteria that no model ever solves are recovered by the solver at 93.1\% and 70.8\%, and a third is recovered by a separate blind decoder on 34 of 40 problems, so those quantities are measurable from the released data rather than defective.
The L5 headroom is a limitation of current models and not of the task, and the solver's 75.7 establishes a lower bound on what is attainable from the released data.

\subsection{Capability Profiles Across EMRB}\label{sec:capability_profiles}

RQ2 asks whether EMRB's five levels test separable cognitive demands and where models systematically fail.
Across the 14 models, mean scores decline from 84.9\% on L1 to 72.3\% on L2, 68.7\% on L3, 53.1\% on L4, and 21.2\% on L5.
The broadly monotonic decline supports the intended difficulty progression, while the small difference between L2 and L3 shows that choosing a measurement procedure and applying communication theory impose comparable demands on current models.

Figure~\ref{fig:level_distribution} (Appendix~\ref{app:full_results}) also shows that difficulty is not explained by a single weak model.
On L4, models must maintain a multi-signal inventory, isolate overlapping components, and bind each measurement to the correct emitter before computing interference or link quantities.
The mean falls by 15.6 points from L3 to L4 as these dependencies are introduced.
L5 creates a larger 31.9 point drop because spectrum survey, waveform recovery, coexistence analysis, and system design must be solved as connected parts of one problem.
Even the best L5 score is only 39.8\%, and the full range is 2.4\% to 39.8\%, indicating that system level synthesis remains difficult across model families.

Per-model profiles reveal that EMRB's five levels test separable cognitive demands rather than a single difficulty axis.
GPT-5.5 and Grok-4.5 score higher on L3 than on L2 (93.2\% vs.\ 86.0\% and 92.3\% vs.\ 86.4\%), even though L3 is the harder tier by cross-model mean (68.7\% against 72.3\%).
This inversion exposes a gap between \emph{formula recall} and \emph{engineering judgment}.
Communication-theory equations such as Shannon capacity and $E_b/N_0$ are densely covered in training corpora and can be applied once relevant parameters are given.
L2, in contrast, requires deciding \emph{which} technique to apply to the raw data, a form of meta-reasoning far less represented in text.

Gemini 3.5 Flash provides the sharpest evidence: it scores 91.9\% on L1, collapses to 46.3\% on L2, then recovers to 72.6\% on L3, showing that a model can measure signal properties and apply textbook formulas yet fail precisely at the connective step of selecting and justifying the right methodology.
L4 produces the widest model separation of any level (range 73.1 points, $\sigma{=}17.6$), confirming that maintaining a coherent multi-signal inventory under overlapping spectra is where underlying reasoning capacity is most sharply discriminated.

The capability profiles translate into a concrete boundary for practical use.
The strongest models are already dependable at the early stages of the workflow, producing calibrated measurements and textbook link calculations above 85\%.
The boundary sits at multi-signal analysis: once several emitters share the spectrum, most models fail to keep each measurement attached to the right emitter.
No model can yet be trusted to turn a crowded-band survey into a system design, where the best score is 39.8\%.

\subsection{ReconPilot Performance}\label{sec:reconpilot_results}

Turning to RQ3, we isolate the contribution of workflow structure by evaluating ReconPilot against free form code execution on the same 200 EMRB problems, using three backbones from the middle of the capability range observed in Figure~\ref{fig:main_results}(a): DeepSeek V4 Flash, Gemini 3.5 Flash, and GPT-5.5.
In the free form setting, each model constructs the analysis process directly from the raw I/Q data, while ReconPilot first supplies a consistent reconnaissance context and then focuses the same backbone on task specific measurement and final verification.
For each backbone, the problem set, scoring procedure, and model configuration are held fixed, so the comparison isolates the effect of organizing the analysis process from the effect of the underlying model.

\begin{table}[t]
\caption{Controlled comparison between free form code execution and ReconPilot across three backbones. Bold marks the higher score in each free form/ReconPilot pair.}
\label{tab:reconpilot_all}
\centering
\resizebox{\columnwidth}{!}{%
\begin{tabular}{llrrrrrr}
\toprule
\textbf{Backbone} & \textbf{Method} & \textbf{L1} & \textbf{L2} & \textbf{L3} & \textbf{L4} & \textbf{L5} & \textbf{Avg.} \\
\midrule
\multirow{3}{*}{DeepSeek V4 Flash}
 & Free form   & \textbf{91.2} & 65.1 & 67.5 & 41.8 & 16.5 & 56.4 \\
 & ReconPilot  & 88.2 & \textbf{86.3} & \textbf{84.9} & \textbf{67.8} & \textbf{24.8} & \textbf{70.4} \\
 & Improvement & $-$3.0 & +21.2 & +17.4 & +26.0 & +8.3 & +14.0 \\
\midrule
\multirow{3}{*}{Gemini 3.5 Flash}
 & Free form   & 91.9 & 46.3 & 72.6 & 47.2 & 28.3 & 57.2 \\
 & ReconPilot  & \textbf{94.9} & \textbf{89.2} & \textbf{84.3} & \textbf{72.2} & \textbf{33.1} & \textbf{74.8} \\
 & Improvement & +3.0 & +42.9 & +11.7 & +25.0 & +4.8 & +17.6 \\
\midrule
\multirow{3}{*}{GPT-5.5}
 & Free form   & 95.8 & 86.0 & \textbf{93.2} & 84.8 & 34.5 & 78.9 \\
 & ReconPilot  & \textbf{97.0} & \textbf{87.7} & 90.1 & \textbf{90.5} & \textbf{48.2} & \textbf{82.7} \\
 & Improvement & +1.2 & +1.7 & $-$3.1 & +5.7 & +13.7 & +3.8 \\
\bottomrule
\end{tabular}}
\end{table}

Table~\ref{tab:reconpilot_all} shows that ReconPilot raises the overall score on all three backbones, by 14.0 points for DeepSeek V4 Flash, 17.6 points for Gemini 3.5 Flash, and 3.8 points for GPT-5.5, and improves 13 of the 15 backbone-level combinations tested.
The two exceptions are informative: DeepSeek V4 Flash loses 3.0 points at L1 and GPT-5.5 loses 3.1 points at L3, both cases where the backbone's free form score was already above 91\%.
When headroom is small, the added reconnaissance context has little to fill and the reduced analysis turn budget becomes the binding constraint.

The largest gains cluster on the multi-signal levels: all three backbones gain the most or second most at L4, and Gemini 3.5 Flash's single largest gain, 42.9 points at L2, comes at the level where its free form score of 46.3\% was its weakest result overall.
This pattern indicates that a model benefits most from ReconPilot when a problem requires tracking several signals at once or when unguided exploration would otherwise stall, and that a fixed reconnaissance context can substitute for capability the backbone lacks.
Appendix~\ref{app:responses} presents two paired free form/ReconPilot case studies (Figures~\ref{fig:case_gemini} and~\ref{fig:case_gpt5}) that trace this at the level of individual sub-questions: at L3 a root symbol rate that free form guesses without measuring propagates into three of five sub-questions, and at L5 an inventory that is missing two of six emitters zeroes 43 points of otherwise correct downstream work through prerequisite gating.
\subsection{Stage Ablation}\label{sec:stage_ablation}

To understand each stage's contribution, we enable Stage~1 and Stage~3 independently on Gemini 3.5 Flash while holding Stage~2 fixed as the free form loop.
Configurations with Stage~3 reserve 3 of 15 turns for verification, so any gain is net of fewer analysis turns.

As shown in Table~\ref{tab:stage_ablation_gemini}, each stage independently helps by a similar amount across all levels: reconnaissance alone lifts the average from 57.2 to 68.8 ($+11.6$), verification alone lifts it to 70.2 ($+13.0$), and both together reach 74.8 ($+17.6$).
However, the combination is markedly sub-additive ($+17.6$ vs.\ $+24.6$ if the two gains were independent), indicating that the two stages repair overlapping failures rather than complementary ones.

At L4, reconnaissance alone is the best configuration (77.9), and adding verification costs 5.7 points because the verification turn re-examines a multi-signal answer that was already correct, consuming analysis turns that L4's complex multi-emitter problems can still use productively.
At L5 the pattern inverts sharply: reconnaissance without verification scores 11.0, which is 17.3 points \emph{below} plain free form, because the unresolved region map invites the model to enumerate sources it cannot separate and nothing downstream checks the result; verification recovers this by catching implausible inventories before they propagate.

\begin{table}[t]
\caption{Stage~1 $\times$ Stage~3 grid on Gemini 3.5 Flash, with Stage~2 held fixed as the unmodified free form loop. Bold marks the best in each column.}
\label{tab:stage_ablation_gemini}
\centering
\resizebox{\columnwidth}{!}{%
\begin{tabular}{ccrrrrrr}
\toprule
\textbf{Stage 1} & \textbf{Stage 3} & \textbf{L1} & \textbf{L2} & \textbf{L3} & \textbf{L4} & \textbf{L5} & \textbf{Avg.} \\
\midrule
--          & --          & 91.9 & 46.3 & 72.6 & 47.2 & 28.3 & 57.2 \\
\checkmark  & --          & 94.8 & 87.8 & 72.6 & \textbf{77.9} & 11.0 & 68.8 \\
--          & \checkmark  & 94.7 & 83.1 & 76.4 & 61.3 & \textbf{35.2} & 70.2 \\
\checkmark  & \checkmark  & \textbf{94.9} & \textbf{89.2} & \textbf{84.3} & 72.2 & 33.1 & \textbf{74.8} \\
\bottomrule
\end{tabular}}
\end{table}

\subsection{Failure Case Analysis}\label{sec:errors}

Returning to the second part of RQ2, the per-criterion scores reveal a progression in why models fail, not just how often.
At \textbf{L1}, individual measurements are usually close to ground truth, but models lose credit by failing to bind each measurement to the correct signal entity.
A model may report accurate frequencies and powers yet assign them to the wrong emitters, suggesting that LLMs treat spectral peaks as isolated numbers rather than maintaining a structured scene inventory.
\textbf{L2} shifts the bottleneck from measurement to justification: models can invoke the right spectral tool but cannot articulate the underlying tradeoff (e.g., sidelobe suppression vs.\ main-lobe widening), indicating procedural fluency without conceptual grounding.

At \textbf{L3}, errors become multiplicative.
A wrong modulation identification cascades through symbol rate, bit rate, and $E_b/N_0$, yet models rarely revisit earlier steps when downstream values become physically implausible, revealing absent self-consistency checking across dependent calculations.
\textbf{L4} compounds this with multi-entity tracking: each measurement must stay bound to its emitter across overlapping spectra, and models that lose binding produce internally coherent but factually wrong answers.

\textbf{L5} represents the full failure mode, where a missed signal in the initial catalog propagates through interference mitigation, waveform recovery, and system design, potentially zeroing an entire question block.
An audit of the extraction-filter sub-question (527 of 560 attempts score zero) separates the collapse into three independent deficits: a perceptual limit where overlapping emitters are merged into one, constraint omission where models solve the unconstrained problem, and absent self-evaluation where models do not check whether their design meets stated requirements.
Detailed per-level case analyses and statistics are provided in Appendix~\ref{app:failure_cases}.

Across all five levels, the pattern is consistent: EMRB failures are rarely about not knowing the right formula.
They arise from the inability to maintain a coherent, entity-bound chain of evidence from raw samples to final answers.

\section{Conclusion}\label{sec:conclusion}

We introduced EMRB, a benchmark for evaluating whether LLMs can analyze raw electromagnetic measurements through executable code.
Scores across 14 LLMs range from 24.1\% to 78.9\% and fall from 84.9\% on basic measurement to 21.2\% on system design, as models struggle to maintain an entity-bound analysis state when measurements become mutually dependent.
ReconPilot raises the overall score by up to 17.6 points, showing that fixed reconnaissance can substitute for missing capability.
EMRB uses only synthetic signals and deterministic scoring, so it can serve as a reproducible acceptance test as models improve.

\section{Limitations}\label{sec:limitations}

EMRB is a controlled synthetic benchmark covering 11 signal types, so its results do not establish model performance on real field measurements with hardware impairments, multipath, or unmodeled signal types; its scope is also narrower than the name suggests, covering RF communication waveforms and LFM radar sweeps under AWGN rather than the full breadth of electromagnetic phenomena.
ReconPilot depends on a fixed PSD-based reconnaissance that may miss weak or overlapping signals, and L5 scores remain low for every backbone, showing that a stable inventory alone does not resolve system design.
We have no human expert baseline; reachability is established by oracle replay, a model union, and a programmatic reference solver (Section~\ref{sec:headroom}), which bounds what is attainable from below.

\section{Ethics and Broader Impact}\label{sec:ethics}

All captures are programmatically synthesized; the dataset contains no real radio recordings, no personal data, and no material requiring ethics review.
Raw spectrum analysis has dual-use relevance to electronic warfare, but EMRB contains no parameters of any deployed system, and its signal models are textbook constructions; we judge the marginal uplift to an adversary to be negligible.

\clearpage
\bibliographystyle{ACM-Reference-Format}
\bibliography{references}

\clearpage
\appendix

\onecolumn

\newcommand{\tocentry}[3]{%
  \noindent\makebox[1.8em][l]{\bfseries#1}#2\nobreak\enspace
  \xleaders\hbox{\kern1.5pt.\kern1.5pt}\hfill\nobreak
  \enspace\makebox[1.6em][r]{#3}\par\vspace{5pt}}

\noindent{\sffamily\bfseries\Large Appendix}\par
\vspace{12pt}

\tocentry{A}{Datasheet for EMRB}{\pageref{app:datasheet}}
\tocentry{B}{Per-Level Problem Examples}{\pageref{app:examples}}
\tocentry{C}{Complete Problem Instance}{\pageref{app:full_problem}}
\tocentry{D}{Evaluation Prompt Templates}{\pageref{app:prompts}}
\tocentry{E}{ReconPilot Reconnaissance Details}{\pageref{app:recon}}
\tocentry{F}{Deterministic Scoring Criteria}{\pageref{app:rubrics}}
\tocentry{G}{Scoring Reliability Analysis}{\pageref{app:scoring_reliability}}
\tocentry{H}{Complete Per-Model Results}{\pageref{app:full_results}}
\tocentry{I}{Results by Question Type}{\pageref{app:question_types}}
\tocentry{J}{Per-Level Failure Case Analysis}{\pageref{app:failure_cases}}
\tocentry{K}{Model Response Examples}{\pageref{app:responses}}
\tocentry{L}{Benchmark Comparison}{\pageref{app:benchmark_table}}

\vfill

\twocolumn
\section{Datasheet for EMRB}\label{app:datasheet}

Following the Datasheets for Datasets framework~\cite{gebru2021datasheets}:

\textbf{Motivation.}
EMRB was created to evaluate LLMs' ability to analyze raw electromagnetic signals through code execution, a capability untested by existing benchmarks.
The benchmark was designed by domain experts in signal processing and wireless communications.

\textbf{Composition.}
200 problems organized into 5~levels $\times$ 40~problems each.
Each problem consists of: (1)~an English-language question with 5~sub-questions (L5:~3), (2)~a complex I/Q signal file (\texttt{.npy}, complex64), and (3)~JSON metadata with generation parameters and ground-truth answers.
The dataset contains 11~modulation types (BPSK, QPSK, 8PSK, 16QAM, 64QAM, 2FSK, 4FSK, FM, AM-DSB, OFDM, Chirp).
SNR ranges from 5\,dB to 30\,dB.
Signal lengths: 32,768 samples (L1--L4) and 65,536 samples (L5).

\textbf{Signal type design rationale.}
The 11 signal types are chosen to cover the major spectral and structural families that appear in real RF environments:
\begin{itemize}[leftmargin=*,nosep]
  \item \textbf{Digital modulations (7):} BPSK, QPSK, 8PSK, 16QAM, 64QAM, 2FSK, 4FSK.
    These span phase-shift keying, quadrature amplitude modulation, and frequency-shift keying.
    Parameters include carrier frequency, symbol rate ($R_s$), roll-off factor ($\alpha$), power, and SNR.
    Low-order formats (BPSK, QPSK) test basic detection and measurement; high-order formats (64QAM) demand reasoning about noise sensitivity and spectral efficiency.
  \item \textbf{Analog modulations (2):} FM and AM-DSB.
    FM requires reasoning about frequency deviation and Carson bandwidth ($B = 2(\Delta f + f_m)$), while AM-DSB exposes carrier-sideband structure and modulation depth.
    Including analog signals prevents models from assuming all waveforms follow digital burst patterns.
  \item \textbf{Wideband signals (2):} OFDM and linear frequency-modulated (LFM) chirp.
    OFDM introduces subcarrier structure, cyclic prefix, and PAPR, enabling questions about multi-carrier system design.
    LFM chirps introduce sweep bandwidth and time-bandwidth product analysis, extending coverage to radar-type waveforms.
\end{itemize}

\textbf{Generation pipeline.}
EMRB generates each problem through a deterministic pipeline that couples signal synthesis with question and answer construction.
The core abstraction is an \emph{(archetype, seed)} pair: each difficulty level defines eight scenario archetypes that fix the signal composition and question selection, and each archetype is instantiated five times with different random seeds, yielding 40 problems per level.
Because the 27 question types are deterministic templates, an (archetype, seed) pair fully determines the I/Q waveform, the question text, and the ground-truth answers; changing only the seed produces a structurally identical but numerically distinct problem.
Each instance proceeds through four stages:
(1)~signal types and parameters are sampled from level-specific ranges;
(2)~complex baseband I/Q samples are synthesized, shifted to assigned carrier frequencies, and combined with calibrated AWGN;
(3)~for L4 and L5, multiple emitters are superimposed into a shared capture with independently assigned frequencies, powers, and SNRs;
(4)~the same parameter record instantiates the question text and computes ground-truth answers analytically or via reference signal-processing code.
The final release stores each problem as a NumPy \texttt{.npy} file (raw I/Q) paired with a \texttt{.json} file (metadata, questions, and ground-truth answers).
All random choices derive from the seed, so the released 200 problems are deterministically reproducible, and the seed space extends well beyond the current dataset: running the same generators with unused seeds produces fresh, valid problems on demand without any manual authoring.

\textbf{Collection process.}
All signals are programmatically generated using the custom signal library released with the benchmark.
Each generator implements standard modulation/signal models with configurable parameters.
Questions are hand-crafted templates instantiated with problem-specific parameters.
Ground truth is computed analytically from generation parameters.

\textbf{Preprocessing/cleaning/labeling.}
Waveform implementation tolerance: power level $\pm$0.5\,dB, SNR $\pm$1\,dB, bandwidth $\pm$5\% of the requested value, no numerical artifacts. This bounds how closely a synthesized waveform realizes its generation target; it is not the accuracy at which answers are scored.
All 200 problems pass validation.

\textbf{Detailed validation procedures.}
EMRB applies three post-generation quality checks to ensure that scored answers are recoverable from the released waveforms:
\begin{enumerate}[leftmargin=*,nosep]
  \item \textbf{Ground-truth verification.}
    For each problem, analytic answers are first computed from the generation parameters.
    The saved \texttt{.npy} file is then loaded and reference signal-processing scripts run on the same I/Q array provided to models.
    For spectral questions, scripts estimate PSDs via Welch's method, detect carrier peaks, and measure bandwidth using the definition stated in the question.
    For power and SNR questions, received power and noise are estimated directly from samples.
    For multi-signal questions, the scripts recover the number of emitters and their carrier locations.
    \emph{Ground-truth recoverability.} Measured quantities are compared with the stored answers; all 200 problems pass with $<$1\% numerical error. This is the property that matters for scoring: every graded value is recoverable from the released waveform.
  \item \textbf{Signal integrity checks.}
    Each saved waveform is validated for physical consistency: correct complex array shape, finite sample values, no NaN/overflow/clipping.
    \emph{Waveform implementation tolerance.} Received power must be within $\pm$0.5\,dB of the generation target, SNR within $\pm$1\,dB, and measured bandwidth within $\pm$5\%. These are looser than the recoverability check above because they bound synthesis fidelity rather than answer accuracy.
    These tolerances ensure that model failures reflect analysis errors, not corrupted inputs.
  \item \textbf{Question coverage validation.}
    The dataset distribution is verified against generated metadata.
    Each L1--L3 and L5 question type appears in all 40 problems of its level. L4 draws five of its nine types per problem, giving the distribution below over its 200 sub-questions.

\begin{center}\small
\begin{tabular}{lrr}
\toprule
\textbf{L4 question type} & \textbf{Problems} & \textbf{Share} \\
\midrule
Symbol rate and modulation order & 40 & 20.0\% \\
Spectral-gap link budget         & 37 & 18.5\% \\
FM parameters                    & 30 & 15.0\% \\
Shannon capacity                 & 27 & 13.5\% \\
Chirp and radar                  & 25 & 12.5\% \\
Interference and SIR             & 16 &  8.0\% \\
Burst analysis                   & 15 &  7.5\% \\
OFDM parameters                  &  5 &  2.5\% \\
AM parameters                    &  5 &  2.5\% \\
\bottomrule
\end{tabular}
\end{center}

    The distribution is uneven by construction: a type is instantiated only when the sampled scene contains the signals it requires, so OFDM and AM parameters appear in the five problems whose archetype includes those emitters.
    Parameter ranges vary within each type, and no two problems of the same type share identical parameters.
    Every signal type appears at least 5 times across the dataset, and the seven most common types (BPSK, QPSK, 8PSK, 16QAM, FM, AM-DSB, Chirp) at least 45 times, preventing evaluation collapse onto a narrow set of waveform patterns.
\end{enumerate}

\textbf{Representativeness.}
The 200 captures contain 805 emitters, 2 to 7 per capture, drawn from 11 base signal types that appear as 15 waveform variants once burst-gated forms are counted. Measured over all emitters, the parameter coverage is:

\begin{center}\small
\begin{tabular}{lrr}
\toprule
\textbf{Quantity} & \textbf{Range} & \textbf{Median} \\
\midrule
Emitters per capture      & 2 to 7                & 3 \\
Occupied bandwidth        & 11\,kHz to 5\,MHz     & 540\,kHz \\
Symbol rate (digital)     & 1 to 750\,ksps        & 119\,ksps \\
Received power            & $-$43.9 to $-$22.4\,dBm & $-$32.5\,dBm \\
In-band SNR               & 18.4 to 57.9\,dB      & 33.0\,dB \\
\bottomrule
\end{tabular}
\end{center}

These ranges are set so that every scored quantity is measurable from the capture rather than to imitate any particular deployment: the SNR floor of 18.4\,dB, for instance, keeps modulation recognition and symbol-rate estimation well posed, which a field capture would not guarantee.
We state the exclusions explicitly. The channel is additive white Gaussian noise only, so there is no multipath, no fading, no Doppler beyond the modeled chirp sweep, and no nonstationary interference. Receiver effects are absent: no I/Q imbalance, no phase noise, no oscillator drift, no amplifier nonlinearity, and no quantization beyond the stored \texttt{complex64} format. Emitters outside the 11 modeled families, including spread-spectrum, frequency-hopping, and radar waveforms other than LFM, do not occur.
The consequence is that EMRB measures analysis competence on well-conditioned captures. It bounds field difficulty from below and should not be read as a sample of any real spectral environment.

\textbf{Utility.}
Four properties support the intended use of comparing and diagnosing models.
It discriminates: reported scores span 24.1\% to 78.9\% with no model saturating, and the spread is not driven by one weak system.
It surfaces structure that aggregate accuracy hides, including that L2 methodology questions are harder than L3 communication theory for most models, that absolute power calibration is the weakest measurement type at both L1 and L2 (Appendix~\ref{app:question_types}), and that L5 failures decompose into three separable deficits rather than one (Section~\ref{sec:errors}).
It is actionable: ReconPilot was designed against these diagnoses and improves three backbones by 3.8 to 17.6 points (Section~\ref{sec:reconpilot_results}).
It is reproducible and expandable: scoring is deterministic with no LLM judge, every result carries a scorer version and a task provenance fingerprint, and the seed-based generation pipeline can produce fresh, valid problems from unused seeds without manual authoring, providing a built-in defense against data contamination.

\textbf{Uses.}
Primary: evaluation of LLMs on EM signal analysis.
Secondary: evaluation of structured analysis pipelines for scientific data analysis; acceptance and regression testing of LLM-based tools before their integration into RF analysis workflows.
Not intended for: training signal classifiers, real-world signal detection.

\textbf{Distribution.}
The dataset, the generation scripts, the deterministic verifiers, and the evaluation harness are released at \url{https://github.com/mingxuZhang2/EMRB}.
The dataset is distributed under CC BY 4.0 and the code under the MIT License.
License: CC-BY-4.0 (data), MIT (evaluation code).
Format: NumPy \texttt{.npy} files + JSON metadata.
Total size: ${\sim}$300\,MB (200 signal files + metadata).

\textbf{Maintenance.}
Maintained by the paper authors.
Planned updates: additional signal types, real-world capture extension, difficulty rebalancing based on model capability frontiers.
Version tracking via semantic versioning (v1.0 at release).

\section{Per-Level Problem Examples}\label{app:examples}

\begin{table}[t]
\caption{EMRB difficulty levels with question types. Each level tests distinct skills with zero question-type overlap across levels. Each L1--L3 and L5 problem instantiates all of its level's question types; each L4 problem draws its five sub-questions from the level's nine types.}
\label{tab:levels}
\centering\small
\setlength{\tabcolsep}{3.5pt}
\begin{tabular}{clp{3cm}cc}
\toprule
\textbf{Lvl} & \textbf{Theme} & \textbf{Question Types} & \textbf{Emitters} & \textbf{Qs} \\
\midrule
L1 & Basic Meas. & Detection, Power, Sampling, Noise Floor, Classification & 2--3 & 5 \\
\addlinespace
L2 & Signal Proc. & FFT/Window, Bandwidth, Autocorrelation, STFT, Energy/Power & 3--4 & 5 \\
\addlinespace
L3 & Comm.\ Theory & Bitrate, Eb/N0, PAPR, ADC, DDC & 3--4 & 5 \\
\addlinespace
L4 & Multi-Signal & Symbol Rate/Mod.\ Order, FM, AM, Chirp/Radar, Burst, Spectral Gap, Interference/SIR, OFDM, Capacity & 3--5 & 5 \\
\addlinespace
L5 & System Design & Spectrum Survey, Radar/Coexist, OFDM Design & 6--7 & 3 \\
\bottomrule
\end{tabular}
\end{table}

This section provides complete, representative problem instances for each difficulty level, showing the full question text, expected analysis approach, and scoring criteria.

\subsection{L1 Example: Signal Detection}

\begin{lvlbox}{lvl1frame}{L1: EMRB\_L1\_5010 \,|\, Signal Detection \& Basic Measurement}
{\small
\textbf{Signal:} 2 signals: BPSK at $f_c{=}+2.63$\,MHz ($R_s{=}350$\,ksps, $\alpha{=}0.35$, $-$28.9\,dBm) and FM at $f_c{=}-4.38$\,MHz (deviation 189\,kHz, $-$32.9\,dBm). $f_s{=}20$\,MHz, $N{=}32{,}768$, noise floor $-$54\,dBm/20\,MHz.\par\smallskip
\textbf{Sub-questions:}\par
\quad Q1: Count the signals, estimate center frequencies, identify the strongest/weakest and their power difference. \hfill \textit{[Numerical]}\par
\quad Q2: Estimate per-signal power (dBm), convert to mW, compute total power and SNR. \hfill \textit{[Numerical]}\par
\quad Q3: Derive sampling parameters and FFT frequency resolution. \hfill \textit{[Numerical]}\par
\quad Q4: Estimate the noise floor from signal-free regions. \hfill \textit{[Numerical]}\par
\quad Q5: Classify each signal's modulation type. \hfill \textit{[Categorical]}\par\smallskip
\textbf{Ground truth:} 2 signals; BPSK at $+$2.63\,MHz (BW$\approx$473\,kHz), FM at $-$4.38\,MHz (Carson BW$\approx$591\,kHz); powers $-$28.9 and $-$32.9\,dBm.\par\smallskip
\textbf{Expected approach:} Compute PSD via Welch method, count peaks above noise floor + 10\,dB, integrate band power, and classify based on spectral shape and envelope statistics.
}
\end{lvlbox}

\subsection{L2 Example: Signal-Processing Methodology}

\begin{lvlbox}{lvl2frame}{L2: EMRB\_L2\_6000 \,|\, Signal-Processing Methodology}
{\small
\textbf{Signal:} 3 signals: QPSK at $+2.44$\,MHz ($R_s{=}250$\,ksps, $\alpha{=}0.3$), FM at $-5.36$\,MHz, LFM chirp sweeping $+4.09$ to $+7.71$\,MHz. $f_s{=}20$\,MHz, $N{=}32{,}768$.\par\smallskip
\textbf{Sub-questions:}\par
\quad Q1: FFT frequency resolution; rectangular vs.\ Hamming window trade-off. \hfill \textit{[Numerical + Deterministic]}\par
\quad Q2: 3-dB, 99\%-power, and null-to-null bandwidth of the QPSK signal. \hfill \textit{[Numerical, $\pm$10\%]}\par
\quad Q3: FFT-based autocorrelation; spacing and source of the dominant periodic comb. \hfill \textit{[Numerical]}\par
\quad Q4: STFT/spectrogram morphology of the three signals. \hfill \textit{[Categorical]}\par
\quad Q5: Per-signal energy and energy-per-bit relationships. \hfill \textit{[Numerical]}\par\smallskip
\textbf{Ground truth:} $\Delta f{=}610.35$\,Hz, recommended window: Hamming; QPSK 3-dB BW $\approx$250\,kHz, 99\% BW $\approx$325\,kHz; autocorrelation comb spacing 78.0\,$\mu$s, attributable to the FM modulating tone.\par\smallskip
\textbf{Key challenge:} The same capture requires a different procedure for each sub-question: three bandwidth definitions need three implementations, and the autocorrelation comb must be attributed to the correct emitter.
}
\end{lvlbox}

\subsection{L3 Example: Communication-Theory Analysis}

\begin{lvlbox}{lvl3frame}{L3: EMRB\_L3\_4000 \,|\, From Measurement to Link Quantities}
{\small
\textbf{Signal:} 3 signals: QPSK at $+1.31$\,MHz ($R_s{=}500$\,ksps, $\alpha{=}0.35$), FM at $-4.52$\,MHz, LFM chirp at $+6.30$\,MHz. $f_s{=}20$\,MHz, $N{=}32{,}768$, noise floor $-$52\,dBm/20\,MHz.\par\smallskip
\textbf{Sub-questions:}\par
\quad Q1: Modulation scheme, symbol rate, bitrate, spectral efficiency. \hfill \textit{[Categorical + Numerical]}\par
\quad Q2: In-band noise power, SNR, $E_b/N_0$, link margin. \hfill \textit{[Numerical]}\par
\quad Q3: PAPR of each signal and PA back-off implications. \hfill \textit{[Numerical]}\par
\quad Q4: Inter-signal dynamic range and minimum ADC ENOB. \hfill \textit{[Numerical]}\par
\quad Q5: DDC frequency shift, LPF cutoff, decimation factor. \hfill \textit{[Numerical]}\par\smallskip
\textbf{Ground truth:} QPSK, $R_s{=}500$\,ksps, $R_b{=}1000$\,kbps; SNR\,$=$\,33.4\,dB, $E_b/N_0{=}31.7$\,dB, margin 22.1\,dB; PAPR $\approx$5.0/0.0/0.0\,dB (QPSK/FM/chirp); min.\ ENOB 12; DDC shift $-1.31$\,MHz, LPF cutoff 405\,kHz.\par\smallskip
\textbf{Key challenge:} Each sub-question chains a code-based measurement with a communication-theory calculation, so a wrong modulation order propagates through bitrate, $E_b/N_0$, and link margin.
}
\end{lvlbox}

\subsection{L4 Example: Multi-Signal Parameter Extraction}

\begin{lvlbox}{lvl4frame}{L4: EMRB\_L4\_1000 \,|\, Multi-Signal Analysis}
{\small
\textbf{Signals:} 4 signals: QPSK at $+0.67$\,MHz (350\,ksps, $-$34.5\,dBm), 8PSK at $+1.69$\,MHz (150\,ksps, $-$32.8\,dBm), 16QAM at $-$1.74\,MHz (400\,ksps, $-$34.3\,dBm), LFM chirp sweeping $+4.81$ to $+8.19$\,MHz ($-$33.7\,dBm). $f_s{=}20$\,MHz, $N{=}32{,}768$, noise floor $-$45\,dBm/20\,MHz.\par\smallskip
\textbf{Sub-questions:}\par
\quad Q1: Symbol rates and modulation orders of the two closely spaced digital signals. \hfill \textit{[Numerical + Categorical]}\par
\quad Q2: Chirp time-bandwidth product, processing gain, range resolution. \hfill \textit{[Numerical]}\par
\quad Q3: Maximum 64QAM link deployable in the $-1.46$ to $+0.43$\,MHz spectral gap. \hfill \textit{[Numerical]}\par
\quad Q4: Most spectrally critical signal pair and their separation/overlap. \hfill \textit{[Deterministic]}\par
\quad Q5: SNR and Shannon capacity of the 8PSK signal. \hfill \textit{[Numerical]}\par\smallskip
\textbf{Key challenge:} Jointly analyzing 4 emitters. Every sub-question depends on a correct signal inventory, and each measurement must stay bound to the right emitter.
}
\end{lvlbox}

\subsection{L5 Example: Spectrum Survey and System Design}

\begin{lvlbox}{lvl5frame}{L5: EMRB\_L5\_2002 \,|\, Integrated Spectrum \& System Analysis}
{\small
\textbf{Signals:} 6 signals in a 20\,MHz capture, including overlapping digital signals, an LFM chirp, an FM signal, and a burst. $f_s{=}20$\,MHz, $N{=}65{,}536$.\par\smallskip
\textbf{Sub-questions:}\par
\quad Q1 (34\,pts): Build a complete signal inventory, identify the largest non-chirp overlap, design a verified extraction passband, and pack guarded channels into the remaining spectrum. \hfill \textit{[Deterministic]}\par
\quad Q2 (33\,pts): Characterize the chirp, recover the overlapped communication waveform, determine its crossing interval, and recover a symbol sequence. \hfill \textit{[Deterministic]}\par
\quad Q3 (33\,pts): Compare link capacity and spectral efficiency, optimize power allocation, and propose a feasible OFDM link. \hfill \textit{[Constraint verification]}\par\smallskip
\textbf{Key challenge:} Measurements from the signal inventory are reused by interference analysis, waveform recovery, and system design. Early errors propagate across the entire problem.
}
\end{lvlbox}

\section{Complete Problem Instance}\label{app:full_problem}

To illustrate the exact input models receive, the box below shows the complete prompt for an L1 problem.
The prompt identifies the signal file, sampling parameters, and recording metadata, followed by five sub-questions with point allocations.
The prior each level hands the model differs, and the difference is deliberate, so we state it explicitly rather than leaving it to be inferred from the example.
L1, L2, and L3 open with a preliminary scan summary (all 40 problems at each of the three levels): it gives the number of independent sources in the band and, for each source, an approximate center frequency (or sweep range, for a chirp) together with a coarse family label such as digitally modulated, analog FM, or linear frequency modulated.
L4 gives no band-level inventory; individual sub-questions point at a spectral region (e.g., ``two digitally modulated signals in the $-0.3$ to $+2.7$\,MHz region'') without stating what else the capture contains.
L5 gives no prior at all beyond the sampling parameters and a warning that overlapping, burst, and radar emissions may be present; the model must build the source list itself, which is what Q1(a) is scored on.
We state the consequences of the L1 prior and of the hint formulas explicitly.
Across the 40 L1 problems the listed frequencies agree with the ground truth to within 0.05\,MHz, well inside the $\pm 0.5$\,MHz that L1 Q1(a) and Q1(b) accept for full credit, so those two sub-questions, 12 of the 100 points at L1, can in principle be answered by restating the prompt rather than by measuring.
A further 44 points ask for arithmetic on quantities the prompt itself supplies or for applying its labels: the Q3 sampling-parameter derivations from the stated $f_s$ and $N$ (20 points), the Q4(a)(b) noise calculations from the stated noise floor (11), the Q4(d) conceptual question (4), and the Q5(c)(d) classifications from the prior's family labels and the hint's envelope rule (9); 5 more points ask a property that is true by construction in every problem.
This is deliberate rather than an oversight: these items test whether a model maintains unit discipline and dB arithmetic before any measurement begins, and weaker models fail exactly there, with Llama 3.3 70B earning 72\% of these points and Kimi-K2.6 83\%.
The remaining 39 points, the per-signal powers, SNRs, and bandwidths of Q1(c)(d), Q2, and Q5(a)(b), are recoverable only from the samples, and this measurement core is where L1 discriminates: scores on it span 18.7\% to 92.9\% across the 14 models, a wider spread than the 52.5\% to 95.8\% of the aggregate.
The L1 aggregate should therefore be read as guided arithmetic plus measurement accuracy against a known scene, and not as evidence of detection ability.
The anchor exists because L1 is meant to isolate measurement: a model that mislocates a source would otherwise forfeit downstream measurement credit for a reason unrelated to measurement, and the level would stop being diagnostic.
From L4 onward the anchor is withdrawn and source enumeration becomes a scored quantity in its own right.

\begin{figure*}[t]
\begin{promptbox}[EMRB\_L1\_5000 \,|\, Complete Problem Prompt (L1: Basic Measurement)]
{\small\ttfamily
You are an electromagnetic signal analysis expert. Below is I/Q signal data collected from an electromagnetic environment.\par\medskip
Signal file: EMRB\_L1\_5000.npy\quad|\quad Sampling rate: 20 MHz\quad|\quad Number of samples: 32768\par
Data format: complex64 (numpy)\quad|\quad Recording duration: 1.6384 ms\par\medskip
This frequency band contains 3 non-overlapping independent signal sources (types include Digital Modulation/Analog FM/Swept Frequency), distributed as follows based on a preliminary scan:\par
\quad\textbullet\ Approx.\ +3.5 MHz: Digital Modulation signal\par
\quad\textbullet\ Approx.\ $-$5.8 MHz: Analog FM signal\par
\quad\textbullet\ Approx.\ $-$9.0\textasciitilde$-$5.8 MHz: Swept Frequency signal (linear frequency modulation)\par\medskip
Please analyze these signals and answer the following questions (20 pts each, 100 pts total):\par\medskip
\textbf{Q1.} Observe the spectrum of this frequency band and answer the following questions:\par
\quad (a) How many distinct independent signals are present in this band? \hfill\textit{(4 pts)}\par
\quad (b) Estimate the center frequency (MHz) of each signal, listed from lowest to highest. \hfill\textit{(8 pts)}\par
\quad (c) Which signal has the strongest power? Which has the weakest? (Identify by center frequency) \hfill\textit{(4 pts)}\par
\quad (d) What is the power difference in dB between the strongest and weakest signals? \hfill\textit{(4 pts)}\par\medskip
\textbf{Q2.} Power measurement and dB conversion:\par
\quad (a) Estimate the power (dBm) of each signal. \hfill\textit{(6 pts)}\par
\quad (b) Convert the power of each signal from dBm to mW.\par
\quad\quad Hint: P(mW) = 10\textasciicircum(P(dBm)/10) \hfill\textit{(4 pts)}\par
\quad (c) Calculate the total power of all signals (dBm).\par
\quad\quad Hint: P\_total = 10$\cdot$log$_{10}$($\Sigma$P$_i$(mW)) \hfill\textit{(4 pts)}\par
\quad (d) Given a noise floor of $-$53.0 dBm/20MHz,\par
\quad\quad calculate the SNR (dB) of each signal within its respective bandwidth.\par
\quad\quad Hint: N\_sig = N$_0$(dBm/Hz) + 10$\cdot$log$_{10}$(BW\_Hz) \hfill\textit{(6 pts)}\par\medskip
\textbf{Q3.} Sampling parameters and frequency resolution:\par
\quad (a) Given sampling rate fs = 20 MHz and number of samples N = 32768,\par
\quad\quad calculate the recording duration T = N/fs (ms). \hfill\textit{(4 pts)}\par
\quad (b) Calculate the sampling interval Ts = 1/fs (ns). \hfill\textit{(4 pts)}\par
\quad (c) Calculate the FFT frequency resolution $\Delta$f = fs/N (Hz). \hfill\textit{(4 pts)}\par
\quad (d) To halve the frequency resolution $\Delta$f, how many samples are needed?\par
\quad\quad Should the sampling rate fs or the number of samples N be changed? Explain your reasoning. \hfill\textit{(8 pts)}\par\medskip
\textbf{Q4.} Noise floor estimation:\par
\quad (a) Observe the signal-free regions of the spectrum and estimate the noise power spectral density (dBm/Hz).\par
\quad\quad Hint: Noise floor (dBm/20MHz) = $-$53.0 dBm,\par
\quad\quad N$_0$(dBm/Hz) = Noise floor (dBm) $-$ 10$\cdot$log$_{10}$(fs) \hfill\textit{(6 pts)}\par
\quad (b) Calculate the noise power within a 1 MHz bandwidth (dBm).\par
\quad\quad Hint: P\_noise = N$_0$(dBm/Hz) + 10$\cdot$log$_{10}$(1$\times$10\textasciicircum6) \hfill\textit{(5 pts)}\par
\quad (c) Determine whether this is white noise (i.e., whether the PSD is approximately uniform\par
\quad\quad across all frequencies). \hfill\textit{(5 pts)}\par
\quad (d) If all signals were removed, would the noise floor change? Why or why not? \hfill\textit{(4 pts)}\par\medskip
\textbf{Q5.} Signal feature classification:\par
\quad (a) Classify each signal by bandwidth:\par
\quad\quad Narrowband ($<$100kHz), Midband (100kHz-1MHz), Wideband ($>$1MHz). \hfill\textit{(5 pts)}\par
\quad (b) Estimate the approximate 3 dB bandwidth (kHz) of each signal. \hfill\textit{(6 pts)}\par
\quad (c) Which signals are constant-envelope signals?\par
\quad\quad Hint: FM and Chirp signals are constant-envelope; PSK modulation depends on implementation. \hfill\textit{(5 pts)}\par
\quad (d) Identify the type of each signal: CW / FM / AM / Digital Modulation / Chirp. \hfill\textit{(4 pts)}\par\medskip
Please provide complete calculation procedures and numerical results for each question.
}
\end{promptbox}
\vspace{-4pt}
\captionof{figure}{Complete L1 problem prompt as delivered to the model, reproduced verbatim from \texttt{EMRB\_L1\_5000.json} with no abridgement, including the hint formulas the prompt supplies. The model receives this text together with an \texttt{execute\_python} tool and the raw I/Q file.}
\label{fig:full_prompt}
\end{figure*}

\section{Evaluation Prompt Templates}\label{app:prompts}

\subsection{System Prompt (Free-Form)}

\begin{codeprompt}[System Prompt]
You are an expert in electromagnetic signal analysis.
You have access to a Python code execution tool.

Your task: analyze the I/Q signal data file and
answer the questions below. The signal is stored as
a complex numpy array (.npy file).

Available libraries: numpy, scipy, matplotlib.
Execution timeout: 30 seconds per call.
Output limit: 10,000 characters per call.

When ready, provide your final answers in this
EXACT format:
===ANSWERS===
Q1a: [your answer]
Q1b: [your answer]
...
===END===

Signal file: {signal_path}
Sampling rate: {fs} Hz
Number of samples: {N}

--- QUESTIONS ---
{question_text}
\end{codeprompt}

\subsection{Stage 2 Input (ReconPilot)}

The reported ReconPilot configuration does not give Stage~2 its own system
prompt. It reuses the free-form system prompt above verbatim, so the only
difference from the free-form baseline is that the region map is prepended to
the question and that three of the fifteen turns are reserved for Stage~3,
leaving Stage~2 twelve turns. The prepended block is:

\begin{codeprompt}[Stage 2 Prepended Block]
## Automated Reconnaissance (localization only; an
ROI is not a signal and ROI count is not source
count)
```
{region_map}
```

{question_text}
\end{codeprompt}

Holding the system prompt fixed is deliberate: it keeps the free-form and
ReconPilot conditions identical except for the reconnaissance context and the
reserved verification turns, so the comparison in
Table~\ref{tab:reconpilot_all} is not confounded by prompt engineering.

\subsection{Force-Answer and Self-Verification Prompts}

\begin{codeprompt}[Force-Answer Prompt]
You have used all available analysis turns.
Based on your analysis so far, provide your FINAL
answers now in the required format:
===ANSWERS===
Q1a: [answer]
...
===END===

If you are unsure about an answer, provide your
best estimate with a brief note.
\end{codeprompt}

\begin{codeprompt}[Stage 3: Self-Verification Prompt]
Review your answers before submission. Check:
1. All sub-questions are answered
2. Numerical values have correct units
3. Values are physically reasonable
4. Format matches ===ANSWERS=== / ===END===

If you find any issues, correct them.
Output your final verified answers now.
\end{codeprompt}

\section{ReconPilot Reconnaissance Details}\label{app:recon}

The Stage~1 reconnaissance script produces an unresolved spectral region map from the raw I/Q file.
Each component and every threshold is listed below; the values quoted are the ones in the released script.

\begin{itemize}[leftmargin=*,nosep]
  \item \textbf{Welch power spectral density.}
    Two-sided density with segment length $\min(4096, N/4)$ and 50\% overlap, converted to dBm/Hz.
    Averaging across segments suppresses sample level noise while preserving spectral structure.
  \item \textbf{Noise-floor estimation.}
    The median of the resulting density. All detection thresholds and reported peak levels are relative to this value, which makes the map's decisions independent of absolute receiver gain.
  \item \textbf{Region detection.}
    The density is smoothed with a 21-bin moving average and thresholded at 6\,dB above the noise floor; a 9-bin binary closing joins fragments of one occupancy, and connected components become candidate regions.
    A region is discarded if it spans fewer than three frequency bins or if its in-region peak is below 10\,dB above the floor, which removes ripple and isolated spurs.
  \item \textbf{Per-region reporting.}
    For each surviving region the script reports the lower and upper frequency limits, the width, the power integrated over the region, and the in-region peak level.
    These are bounds on where energy is, not per-source measurements: a region is not assumed to hold exactly one emitter, so no center frequency, 3-dB bandwidth, or per-source SNR is claimed.
  \item \textbf{Short-time Fourier observations.}
    A two-sided STFT with segment length $\min(1024, N/8)$ supplies three statistics per region: the 90th minus 10th percentile spread of total region energy over time, the 95th percentile of the same spread computed within individual frequency bins, and the 5th to 95th percentile span of the dominant frequency.
    These expose burst behavior and frequency drift that a single averaged density would hide.
  \item \textbf{Qualitative flags.}
    The statistics are reduced to flags rather than parameter estimates: \emph{broad} above 1\,MHz of width, \emph{time-varying} above 6\,dB of total spread, \emph{locally-time-varying} above 8\,dB of within-bin spread, \emph{moving-dominant-frequency} when the dominant frequency span exceeds $\max(0.35\,\text{MHz}, 0.45 \times \text{width})$, and \emph{unresolved-stationary} when none of these apply.
  \item \textbf{Follow-up instruction.}
    The map closes by telling the model to audit the source count in every region, to inspect the time frequency ridge and the residual after masking it for broad or moving regions, to compare active against inactive spectra per frequency bin for time varying regions, and not to assign source identities from the map alone.
\end{itemize}

The script does not count sources, classify modulations, estimate symbol rates, or fit chirp parameters, and it computes no time-domain amplitude statistics and no autocorrelation.
Its header states explicitly that a region may contain one source, several overlapping sources, a burst, or a swept signal, and that the region count is not the source count.
This is a deliberate limit rather than an omission: a fixed script cannot separate overlapping emitters reliably, and a confidently wrong inventory propagates into every later measurement, whereas an acknowledged bound does not.
The output is plain text, prepended to the Stage~2 prompt (see Appendix~\ref{app:prompts}), and is deterministic and model independent: every backbone receives the same map for a given problem.

\section{Deterministic Scoring Criteria}\label{app:rubrics}

Numerical scoring is defined separately for each physical quantity rather than by one global error formula.
Each criterion specifies accepted units, an absolute or relative tolerance, and the signal or sub-question to which the value must be bound.
This distinction is necessary because a frequency error is naturally measured in hertz, a power error in decibels, and a signal inventory must match several identities rather than one scalar target.

Structured answers are verified by evaluating their consequences.
For example, an L5 extraction passband must retain the required fraction of the target bandwidth and produce the reported SIR under the stated attenuation model.
Channel placements are checked against band edges, occupied intervals, and guard constraints, while OFDM proposals are checked for bandwidth, power, leakage, and rate feasibility.
Waveform recovery accepts the documented phase, scale, and conjugation equivalences before measuring symbol agreement.
These verifiers allow multiple valid numerical designs without delegating correctness to an LLM judge.

\subsection{Categorical Answer Matching}

The auto-scorer implements alias resolution for modulation type matching:

\begin{table}[h]
\centering\small
\begin{tabular}{ll}
\toprule
\textbf{Ground Truth} & \textbf{Accepted Aliases} \\
\midrule
QPSK & 4-QAM, 4QAM, Quadrature PSK, 4-PSK \\
16QAM & 16-QAM, QAM-16, QAM16 \\
FM & Frequency Modulation, FM modulation \\
AM-DSB & AM, DSB-AM, Double-Sideband AM \\
OFDM & Orthogonal FDM \\
Chirp & LFM, Linear FM, Swept frequency \\
\bottomrule
\end{tabular}
\end{table}

\subsection{Decision and Rationale Criteria}\label{app:rationale_criteria}

Some prompts ask for a decision rather than a value: whether the sampling rate or the sample count should change (L1), which window or bandwidth definition to use (L2), or whether a link closes (L3).
The verifiers score these with negation-aware assertion checks: the conclusion must be asserted, not merely mentioned, so a response stating ``do not increase $N$'' names the expected vocabulary while asserting the opposite and scores zero.
Prompts that additionally say ``explain your reasoning'' are, with the four exceptions below, scored on the conclusion alone, and the explanation itself earns no credit.
L1 Q3(d), for example, awards 4 points for the sample count that halves the resolution and 4 points for asserting that the sample count rather than the sampling rate must grow.

Four L2 criteria go further and credit the content of the stated rationale.
Each decomposes the expected argument into components that can be checked independently, and a component is credited only when the response asserts it with the correct direction near the relevant concept term.
The window-selection rationale, for instance, is split into three directional claims, namely that a Hamming window lowers sidelobes, widens the main lobe, and reduces spectral leakage; a claim asserted with the wrong direction, or negated, contributes nothing.
The four criteria and their weights are the window-selection rationale (3 points), the bandwidth-definition distinctions (2), the allocation-planning rationale (1), and the dBm-summation explanation (5, of which one third is a numeric verification against the response's own totals).
They total 11 of the 100 L2 points and 2.2\% of the benchmark; no other level credits rationale content.

We state the limitation plainly.
These checks verify that the correct claims are asserted, not that they were derived, so a response could in principle earn them by stating the right facts without performing the analysis.
The exposure is bounded: earning all 11 points by assertion alone raises an overall score by at most 2.2 points against an observed spread of 54.8 points.
In practice the criteria are among the harder items of their level rather than free credit: the 14 models earn on average 62\% of the rationale points against 74\% of the remaining L2 points, and the deficit is largest for several of the strongest models (GPT-5.5 earns 50\% of the rationale credit and 90\% of the rest), consistent with the finding in Section~\ref{sec:errors} that articulating a tradeoff is harder for current models than executing the measurement.

\section{Scoring Reliability Analysis}\label{app:scoring_reliability}

All reported questions are routed to versioned deterministic verifiers.
Given the same final response, these verifiers produce identical criterion and total scores without an external model call.
The parsers bind quantities to sub-question labels, units, and signal identities, while structured L4 and L5 outputs are checked against explicit physical and design constraints.
Every stored result also carries a scorer version and a task provenance fingerprint.
The reporting scripts accept a model and level only when all 40 expected samples match the current task inventory and scorer version, preventing stale or partial results from entering a figure or aggregate score.

\subsection{Zero-Pass Criterion Audit}

Section~\ref{sec:headroom} reports that six criteria are never scored perfectly by any model across 560 attempts, all of them at L5.
We audited each one to separate genuine difficulty from a defect in the task or the rubric.
Two of them, the achieved spectral efficiency and the channel capacity in Q3(a), are answered by the reference solver at 70.8\% and 93.1\%, so they are measurable from the released waveform.
One, the modulation inventory in Q1(a), averages a partial credit of 60\% under the union and requires every emitter in a band of six or seven signals to be labeled correctly at once, which makes it hard rather than impossible.
Two more, the objective value and the reported metrics of the extraction filter in Q1(c), are constraint satisfaction checks rather than comparisons against a stored answer, and the stored optimal passband itself lands within $2 \times 10^{-4}$ of the declared 0.8 in-band fraction, so designs that target exactly 80\% sit on the constraint boundary.
The last, the 32 recovered symbols in Q2(d), was the strictest case: every one of the 560 attempts scored exactly zero.
An independent decoder that reads only the waveform and the question text recovers the symbol stream perfectly on 34 of 40 problems, confirming the task is solvable, and inspection of the model responses shows that 408 of 503 parsed submissions contain only four distinct values, that is, textbook constellation points supplied in place of a failed recovery.

\section{Complete Per-Model Results}\label{app:full_results}

Table~\ref{tab:full_results} reports the level specific scores used in the main analysis.

\begin{table}[t]
\caption{Complete EMRB results for all 14 evaluated LLMs. The best score in each column is bold. Averages are computed from unrounded per-level scores, so they can differ in the last digit from the mean of the rounded values shown.}
\label{tab:full_results}
\centering
\resizebox{\columnwidth}{!}{%
\begin{tabular}{lrrrrrr}
\toprule
\textbf{Model} & \textbf{L1} & \textbf{L2} & \textbf{L3} & \textbf{L4} & \textbf{L5} & \textbf{Avg.} \\
\midrule
GPT-5.5            & \textbf{95.8} & 86.0 & \textbf{93.2} & \textbf{84.8} & 34.5 & \textbf{78.9} \\
Grok-4.5           & 93.4 & 86.4 & 92.3 & 77.2 & \textbf{39.8} & 77.8 \\
Claude Opus 4.7    & 84.0 & \textbf{88.7} & 88.3 & 70.0 & 20.5 & 70.3 \\
GLM-5.2            & 92.9 & 83.8 & 73.7 & 58.6 & 20.0 & 65.8 \\
Claude Sonnet 4.6  & 91.5 & 82.6 & 80.2 & 58.1 & 15.8 & 65.6 \\
Gemini 3.1 Pro     & 91.7 & 79.6 & 57.8 & 63.4 & 34.8 & 65.5 \\
DeepSeek V4 Pro    & 90.8 & 77.9 & 72.8 & 52.7 & 18.7 & 62.6 \\
Gemini 3.5 Flash   & 91.9 & 46.3 & 72.6 & 47.2 & 28.3 & 57.2 \\
MiniMax-M3         & 87.7 & 62.5 & 71.2 & 43.2 & 19.8 & 56.9 \\
MiMo-V2.5-Pro      & 83.7 & 72.0 & 59.4 & 52.9 & 16.3 & 56.8 \\
DeepSeek V4 Flash  & 91.2 & 65.1 & 67.5 & 41.8 & 16.5 & 56.4 \\
DeepSeek-Chat      & 75.2 & 79.6 & 53.5 & 36.1 & 16.7 & 52.2 \\
Kimi-K2.6          & 66.4 & 79.7 & 48.6 & 45.9 & 12.5 & 50.6 \\
Llama 3.3 70B      & 52.5 & 22.7 & 31.2 & 11.7 & 2.4 & 24.1 \\
\bottomrule
\end{tabular}}
\end{table}

\begin{figure}[t]
  \centering
  \includegraphics[width=\columnwidth]{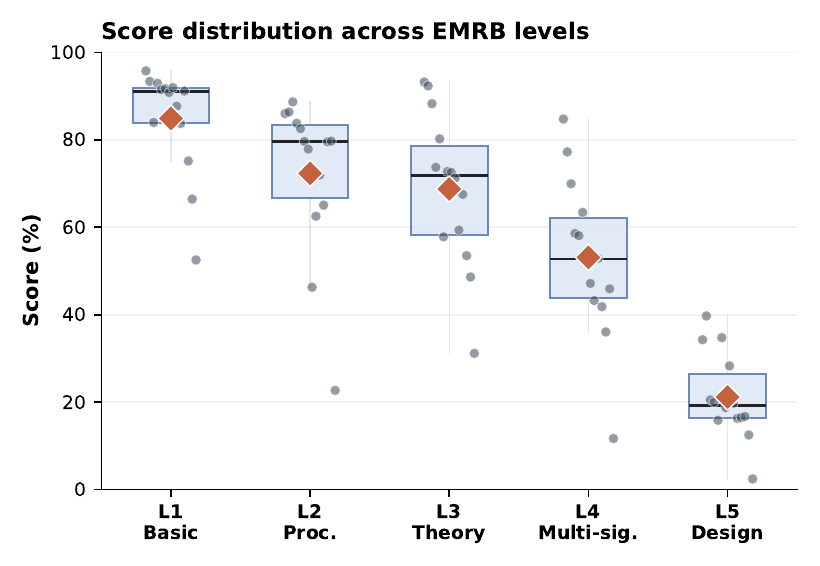}
  \caption{Score distributions across the five EMRB levels. Each gray point represents one model, the black line is the median, and the orange diamond is the mean.}
  \label{fig:level_distribution}
\end{figure}

\section{Results by Question Type}\label{app:question_types}

Figure~\ref{fig:question_types} disaggregates every reported score to the 27 individual question types.
L1, L2, L3, and L5 instantiate all of their types in every problem, so each cell there averages 40 problems; L4 samples 5 of its 9 types per problem, so its cells average between 5 and 40 problems depending on the type.
All values pass the same validation used for the main results.

Three patterns are visible that the level averages hide.
First, the three weakest types in the benchmark are the three L5 types, at 15, 23, and 25 points out of 100 averaged over models, confirming that the L5 deficit is spread across spectrum survey, coexistence analysis, and design rather than concentrated in one sub-question.
Second, Shannon capacity is the weakest L4 type at 43 even though it is the most standard textbook formula in the level, which locates the failure in extracting the right bandwidth and noise terms from a multi-emitter scene rather than in the calculation itself.
Third, absolute power in dBm is the only weak type at L1, at 67 against 81 to 96 for the other four, and the weakest model reaches only 12; calibrated power is therefore the first quantity to break and it stays broken, since the L2 energy and PSD type is also the weakest of its level at 54.

\section{Per-Level Failure Case Analysis}\label{app:failure_cases}

This section provides concrete examples of the failure patterns summarized in Section~\ref{sec:errors}, with one representative case per difficulty level.

\textbf{L1: Signal-identity binding failure.}
On EMRB\_L1\_5000 (3 signals), Llama~3.3~70B correctly detects three emitters but reports the chirp center frequency as $-$9.0\,MHz instead of the ground-truth $-$7.41\,MHz, placing it outside the $\pm$0.5\,MHz full-credit tolerance.
The misplaced chirp forfeits its own slot in the frequency list and then propagates: the strongest and weakest selections, the power difference, the per-signal power and SNR estimates, and the bandwidth category all lose credit, leaving 9.3/20 on Q1 and 5.3/20 on Q2.
GPT-5.5, by contrast, runs calibrated PSD code and reports $-$7.41\,MHz, maintaining correct binding throughout (see Figure~\ref{fig:response_comparison}).

\textbf{L2: Procedural fluency without conceptual grounding.}
On L2 bandwidth-definition problems, models frequently compute all three bandwidth measures (3-dB, null-to-null, 99\%-power) correctly but lose credit on the comparison sub-question by asserting that 3-dB bandwidth is ``most appropriate for spectrum allocation'' without explaining why occupied bandwidth better captures out-of-band energy.
The verifier scores this deterministically: committing to the occupied-power definition earns 2 points, and 1 further point requires asserting its consequence, energy capture or adjacent-channel interference, near that choice; the check verifies which claims are asserted, not the quality of the prose (Appendix~\ref{app:rationale_criteria}).

\textbf{L3: Cascading chain errors.}
On L3 link-budget problems, a model that misidentifies QPSK as BPSK halves the estimated bit rate, which shifts the required $E_b/N_0$ by $\sim$3\,dB, flips the link-margin sign, and produces an incorrect feasibility verdict.
Each of the five chained sub-questions has its own $\pm$3\,dB tolerance, but the propagated error exceeds this at every stage, resulting in 0/100 on an otherwise well-structured analysis.

\textbf{L4: Multi-entity binding collapse.}
On L4 multi-signal problems, models must classify whether two signals overlap or are spectrally separated before computing isolation and interference.
A model that reports the correct occupied bandwidths but swaps which signal occupies which frequency range produces a wrong overlap verdict, which gates 40\% of the sub-question score and invalidates all downstream interference calculations.

\textbf{L5: Catalog-gated scoring cascade.}
On L5 system-design problems, Q1 requires a complete signal inventory matched via a cost-matrix assignment to the reference catalog.
A model that misses one of six signals (e.g., a weak burst emitter 8\,dB below the strongest signal) receives zero on that catalog entry, which then zeros the prerequisite-gated sub-questions in Q2 (interference analysis, 33\,pts) or Q3 (system design, 33\,pts) that depend on the missed signal's identity.

\textbf{L5 extraction-filter deficit breakdown.}
An audit of the extraction-filter sub-question, on which 527 of 560 attempts score zero, accounts for every attempt.
(1)~\emph{Perceptual limit} produces all 527 zeros: 35 submissions contain no schema-valid answer, and 492 name a wrong extraction target.
In 427 of the 492 the true target never appears in the submitted inventory at all, and 390 of these report two reference emitters as a single merged source, at a median center separation of 0.303\,MHz against bandwidths of 0.44 and 0.52\,MHz, with the target 1.3 to 10.6\,dB weaker; in the other 65 the target is inventoried but another emitter is selected.
The remaining 33 attempts name the target correctly: 14 fail the Q1b pair prerequisite, so their designs are never evaluated, and 19 designs reach evaluation.
(2)~\emph{Constraint omission} covers 15 of the 19 evaluated designs: 13 keep the passband on the target but retain only 0.47 to 0.77 of its bandwidth against the required 0.8, 2 place the passband so that almost none of the target falls inside it, and 3 of the 15 reach a post-filter SIR of 30 to 34\,dB, better than the constrained optimum, because they solved the unconstrained problem instead.
(3)~\emph{Absent self-evaluation} covers all 19: every design reports a retained fraction more than 0.03 away from what its own passband delivers, and each of the 15 infeasible designs claims a fraction at or above the 0.8 floor while delivering between 0.00 and 0.77, applying the claimed stopband attenuation to the whole interferer rather than only to the part falling outside the passband.
The 4 remaining designs are feasible, 3 of them within 4\,dB of the optimal SIR, yet no attempt out of 560 earns full credit, because the reported metrics never match the measured behavior of the submitted design.
The second and third deficits are linked, since performing the third check would expose the second, which is why we count self-verification separately from optimization ability.

\section{Model Response Examples}\label{app:responses}

\begin{figure*}[t]
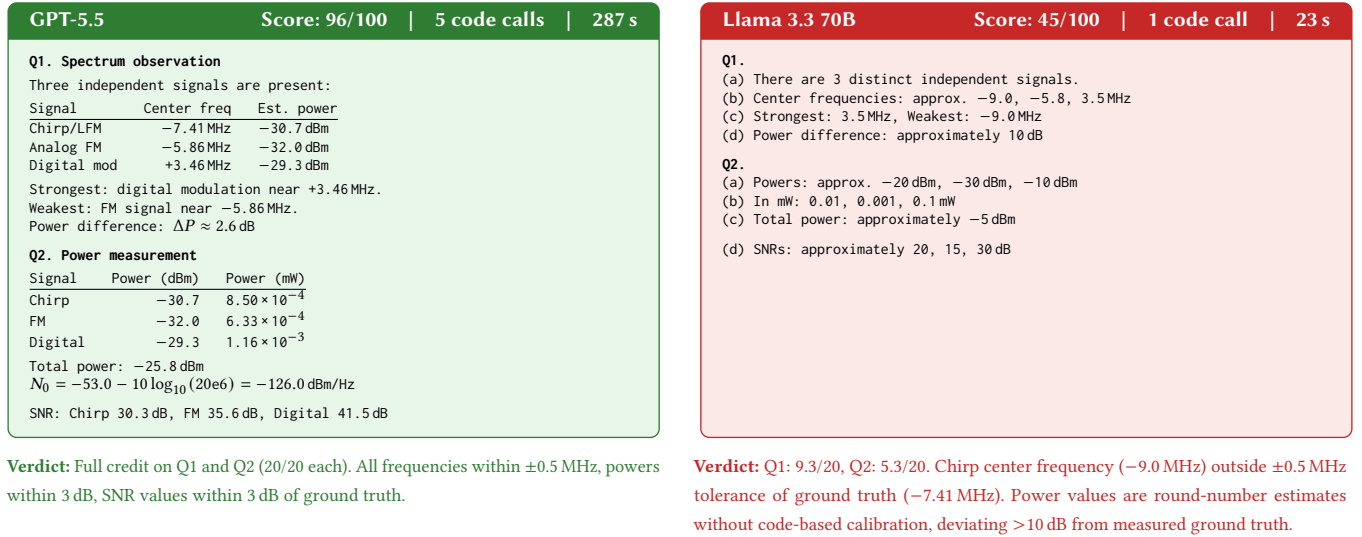

\begin{tcbraster}[raster columns=2, raster equal height=rows,
  raster column skip=0.03\textwidth, raster row skip=0pt]
\begin{goodbox}[\textcolor{white}{GPT-5.5} \hfill Score: 96/100\quad|\quad 5 code calls\quad|\quad 287\,s]
{\scriptsize\ttfamily
\textbf{Q1. Spectrum observation}\par\smallskip
Three independent signals are present:\par\smallskip
\begin{tabular}{@{}lrl@{}}
Signal & Center freq & Est.\ power \\
\hline
Chirp/LFM & $-$7.41\,MHz & $-$30.7\,dBm \\
Analog FM & $-$5.86\,MHz & $-$32.0\,dBm \\
Digital mod & +3.46\,MHz & $-$29.3\,dBm \\
\end{tabular}\par\smallskip
Strongest: digital modulation near +3.46\,MHz.\par
Weakest: FM signal near $-$5.86\,MHz.\par
Power difference: $\Delta P \approx 2.6$\,dB\par\medskip
\textbf{Q2. Power measurement}\par\smallskip
\begin{tabular}{@{}lrr@{}}
Signal & Power (dBm) & Power (mW) \\
\hline
Chirp & $-$30.7 & 8.50\,\texttimes\,10$^{-4}$ \\
FM & $-$32.0 & 6.33\,\texttimes\,10$^{-4}$ \\
Digital & $-$29.3 & 1.16\,\texttimes\,10$^{-3}$ \\
\end{tabular}\par\smallskip
Total power: $-$25.8\,dBm\par
$N_0 = -53.0 - 10\log_{10}(20\text{e}6) = -126.0$\,dBm/Hz\par
SNR: Chirp 30.3\,dB, FM 35.6\,dB, Digital 41.5\,dB
}
\end{goodbox}
\begin{badbox}[\textcolor{white}{Llama 3.3 70B} \hfill Score: 45/100\quad|\quad 1 code call\quad|\quad 23\,s]
{\scriptsize\ttfamily
\textbf{Q1.}\par
(a) There are 3 distinct independent signals.\par
(b) Center frequencies: approx.\ $-$9.0, $-$5.8, 3.5\,MHz\par
(c) Strongest: 3.5\,MHz, Weakest: $-$9.0\,MHz\par
(d) Power difference: approximately 10\,dB\par\medskip
\textbf{Q2.}\par
(a) Powers: approx.\ $-$20\,dBm, $-$30\,dBm, $-$10\,dBm\par
(b) In mW: 0.01, 0.001, 0.1\,mW\par
(c) Total power: approximately $-$5\,dBm\par
(d) SNRs: approximately 20, 15, 30\,dB
}
\end{badbox}
\end{tcbraster}
\vspace{2pt}
\noindent
\begin{minipage}[t]{0.485\textwidth}
{\footnotesize\color{goodframe}\textbf{Verdict:} Full credit on Q1 and Q2 (20/20 each). All frequencies within $\pm$0.5\,MHz, powers within 3\,dB, SNR values within 3\,dB of ground truth.}
\end{minipage}%
\hfill
\begin{minipage}[t]{0.485\textwidth}
{\footnotesize\color{badframe}\textbf{Verdict:} Q1: 9.3/20, Q2: 5.3/20. Chirp center frequency ($-$9.0\,MHz) outside $\pm$0.5\,MHz tolerance of ground truth ($-$7.41\,MHz). Power values are round-number estimates without code-based calibration, deviating $>$10\,dB from measured ground truth.}
\end{minipage}
\vspace{4pt}
\captionof{figure}{L1 response comparison on EMRB\_L1\_5000 (Q1--Q2 excerpts). GPT-5.5 executes multiple rounds of calibrated signal-processing code; Llama~3.3~70B produces superficially plausible answers from the prompt hint without code-verified measurements.}
\label{fig:response_comparison}
\end{figure*}

\begin{figure*}[t]
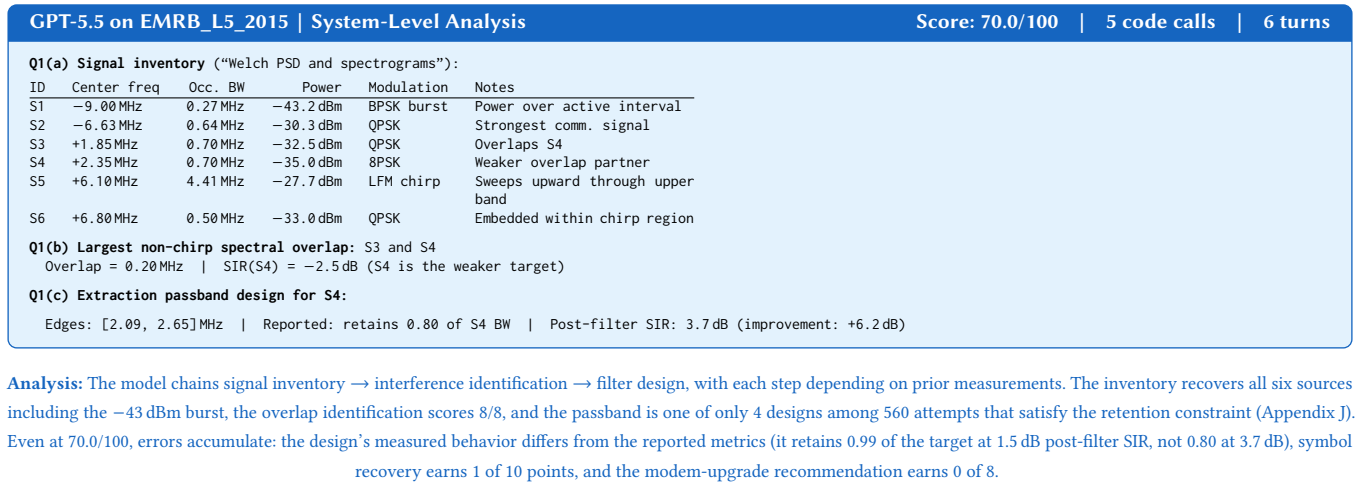

\begin{l5box}[GPT-5.5 on EMRB\_L5\_2015 \,|\, System-Level Analysis \hfill Score: 70.0/100\quad|\quad 5 code calls\quad|\quad 6 turns]
{\scriptsize\ttfamily
\textbf{Q1(a) Signal inventory} (``Welch PSD and spectrograms''):\par\smallskip
\begin{tabular}{@{}clrrlp{2.9cm}@{}}
ID & Center freq & Occ.\ BW & Power & Modulation & Notes \\
\hline
S1 & $-$9.00\,MHz & 0.27\,MHz & $-$43.2\,dBm & BPSK burst & Power over active interval \\
S2 & $-$6.63\,MHz & 0.64\,MHz & $-$30.3\,dBm & QPSK & Strongest comm.\ signal \\
S3 & +1.85\,MHz & 0.70\,MHz & $-$32.5\,dBm & QPSK & Overlaps S4 \\
S4 & +2.35\,MHz & 0.70\,MHz & $-$35.0\,dBm & 8PSK & Weaker overlap partner \\
S5 & +6.10\,MHz & 4.41\,MHz & $-$27.7\,dBm & LFM chirp & Sweeps upward through upper band \\
S6 & +6.80\,MHz & 0.50\,MHz & $-$33.0\,dBm & QPSK & Embedded within chirp region \\
\end{tabular}\par\medskip
\textbf{Q1(b) Largest non-chirp spectral overlap:} S3 and S4\par
\quad Overlap = 0.20\,MHz\quad|\quad SIR(S4) = $-$2.5\,dB (S4 is the weaker target)\par\medskip
\textbf{Q1(c) Extraction passband design for S4:}\par
\quad Edges: [2.09, 2.65]\,MHz\quad|\quad Reported: retains 0.80 of S4 BW\quad|\quad Post-filter SIR: 3.7\,dB (improvement: +6.2\,dB)
}
\end{l5box}
\vspace{2pt}
{\footnotesize\color{l5frame}\textbf{Analysis:} The model chains signal inventory $\to$ interference identification $\to$ filter design, with each step depending on prior measurements. The inventory recovers all six sources including the $-$43\,dBm burst, the overlap identification scores 8/8, and the passband is one of only 4 designs among 560 attempts that satisfy the retention constraint (Appendix~\ref{app:failure_cases}). Even at 70.0/100, errors accumulate: the design's measured behavior differs from the reported metrics (it retains 0.99 of the target at 1.5\,dB post-filter SIR, not 0.80 at 3.7\,dB), symbol recovery earns 1 of 10 points, and the modem-upgrade recommendation earns 0 of 8.}
\vspace{4pt}
\captionof{figure}{L5 system-design response excerpt. EMRB L5 requires chained reasoning where the signal inventory feeds interference analysis, which feeds filter and system design. The verifier checks each step against ground truth and physical constraints.}
\label{fig:l5_response}
\end{figure*}

This section presents actual model responses to illustrate how analysis quality differs across capability levels.
We compare GPT-5.5 and Llama~3.3~70B on the same L1 problem (Figure~\ref{fig:response_comparison}), followed by a GPT-5.5 excerpt on an L5 system-design problem (Figure~\ref{fig:l5_response}).
We then present three paired case studies, one per ReconPilot backbone, contrasting a free-form response with the ReconPilot response on the identical problem to show concretely how the structured pipeline converts a partially-scored free-form attempt into a substantially higher-scoring one.

\begin{figure*}[t]
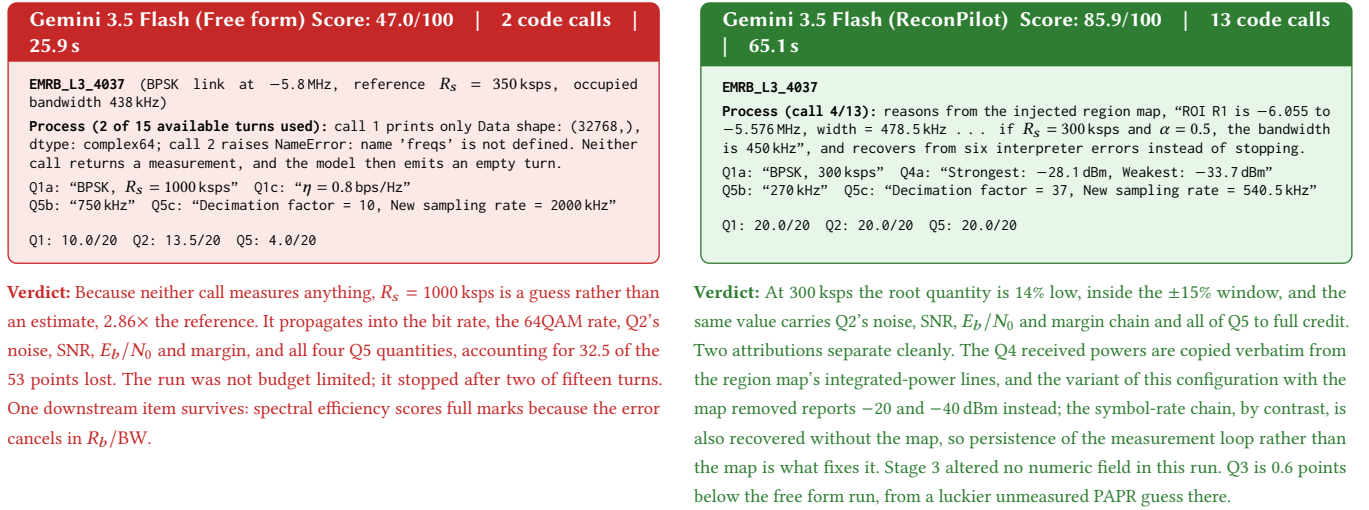

\begin{tcbraster}[raster columns=2, raster equal height=rows,
  raster column skip=0.03\textwidth, raster row skip=0pt]
\begin{badbox}[\textcolor{white}{Gemini 3.5 Flash (Free form)} \hfill Score: 47.0/100\quad|\quad 2 code calls\quad|\quad 25.9\,s]
{\scriptsize\ttfamily
\textbf{EMRB\_L3\_4037} (BPSK link at $-$5.8\,MHz, reference $R_s=350$\,ksps, occupied bandwidth 438\,kHz)\par\smallskip
\textbf{Process (2 of 15 available turns used):} call 1 prints only \texttt{Data shape: (32768,), dtype: complex64}; call 2 raises \texttt{NameError: name 'freqs' is not defined}. Neither call returns a measurement, and the model then emits an empty turn.\par\smallskip
Q1a: ``BPSK, $R_s=1000$\,ksps''\quad Q1c: ``$\eta=0.8$\,bps/Hz''\par
Q5b: ``750\,kHz''\quad Q5c: ``Decimation factor = 10, New sampling rate = 2000\,kHz''\par\smallskip
Q1: 10.0/20\quad Q2: 13.5/20\quad Q5: 4.0/20
}
\end{badbox}
\begin{goodbox}[\textcolor{white}{Gemini 3.5 Flash (ReconPilot)} \hfill Score: 85.9/100\quad|\quad 13 code calls\quad|\quad 65.1\,s]
{\scriptsize\ttfamily
\textbf{EMRB\_L3\_4037}\par\smallskip
\textbf{Process (call 4/13):} reasons from the injected region map, ``\emph{ROI R1 is $-$6.055 to $-$5.576\,MHz, width = 478.5\,kHz \ldots\ if $R_s=300$\,ksps and $\alpha=0.5$, the bandwidth is 450\,kHz}'', and recovers from six interpreter errors instead of stopping.\par\smallskip
Q1a: ``BPSK, 300\,ksps''\quad Q4a: ``Strongest: $-$28.1\,dBm, Weakest: $-$33.7\,dBm''\par
Q5b: ``270\,kHz''\quad Q5c: ``Decimation factor = 37, New sampling rate = 540.5\,kHz''\par\smallskip
Q1: 20.0/20\quad Q2: 20.0/20\quad Q5: 20.0/20
}
\end{goodbox}
\end{tcbraster}
\vspace{2pt}
\noindent
\begin{minipage}[t]{0.485\textwidth}
{\footnotesize\color{badframe}\textbf{Verdict:} Because neither call measures anything, $R_s=1000$\,ksps is a guess rather than an estimate, $2.86\times$ the reference. It propagates into the bit rate, the 64QAM rate, Q2's noise, SNR, $E_b/N_0$ and margin, and all four Q5 quantities, accounting for 32.5 of the 53 points lost. The run was not budget limited; it stopped after two of fifteen turns. One downstream item survives: spectral efficiency scores full marks because the error cancels in $R_b/\mathrm{BW}$.}
\end{minipage}%
\hfill
\begin{minipage}[t]{0.485\textwidth}
{\footnotesize\color{goodframe}\textbf{Verdict:} At 300\,ksps the root quantity is 14\% low, inside the $\pm15\%$ window, and the same value carries Q2's noise, SNR, $E_b/N_0$ and margin chain and all of Q5 to full credit. Two attributions separate cleanly. The Q4 received powers are copied verbatim from the region map's integrated-power lines, and the variant of this configuration with the map removed reports $-$20 and $-$40\,dBm instead; the symbol-rate chain, by contrast, is also recovered without the map, so persistence of the measurement loop rather than the map is what fixes it. Stage~3 altered no numeric field in this run. Q3 is 0.6 points below the free form run, from a luckier unmeasured PAPR guess there.}
\end{minipage}
\vspace{4pt}
\captionof{figure}{Gemini 3.5 Flash on EMRB\_L3\_4037. Free form execution guesses a root symbol rate after two code calls that returned no measurement, and the $2.9\times$ error propagates through Q1's rate quantities, Q2's noise and margin chain, and all of Q5. ReconPilot derives the root quantity from the reconnaissance region map and reaches full credit on Q1, Q2 and Q5.}
\label{fig:case_gemini}
\end{figure*}

\begin{figure*}[t]
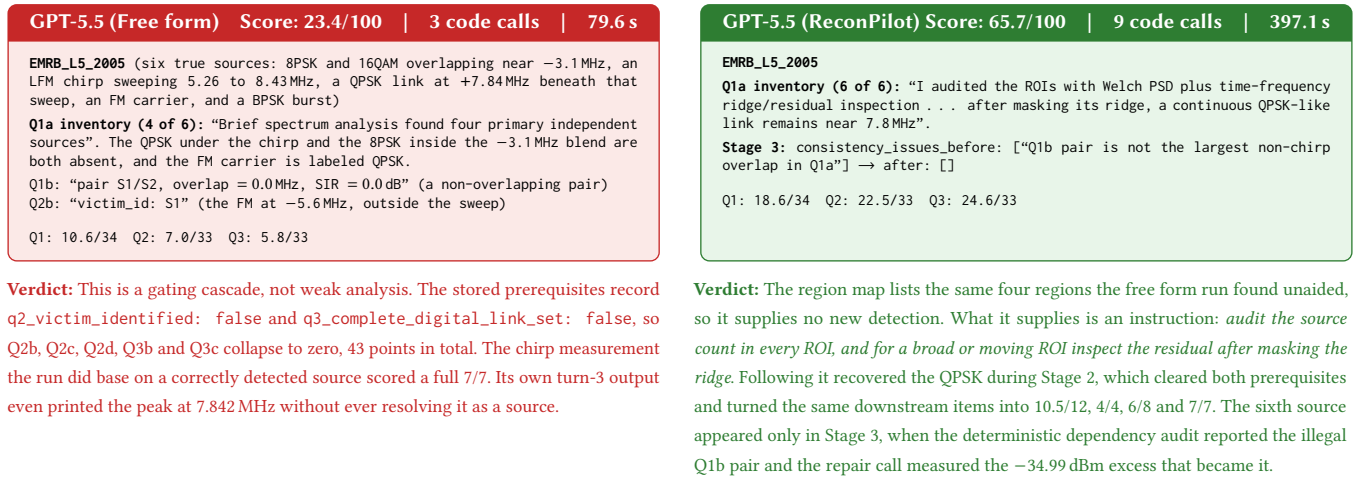

\begin{tcbraster}[raster columns=2, raster equal height=rows,
  raster column skip=0.03\textwidth, raster row skip=0pt]
\begin{badbox}[\textcolor{white}{GPT-5.5 (Free form)} \hfill Score: 23.4/100\quad|\quad 3 code calls\quad|\quad 79.6\,s]
{\scriptsize\ttfamily
\textbf{EMRB\_L5\_2005} (six true sources: 8PSK and 16QAM overlapping near $-$3.1\,MHz, an LFM chirp sweeping 5.26 to 8.43\,MHz, a QPSK link at $+$7.84\,MHz beneath that sweep, an FM carrier, and a BPSK burst)\par\smallskip
\textbf{Q1a inventory (4 of 6):} ``\emph{Brief spectrum analysis found four primary independent sources}''. The QPSK under the chirp and the 8PSK inside the $-$3.1\,MHz blend are both absent, and the FM carrier is labeled QPSK.\par\smallskip
Q1b: ``pair S1/S2, overlap $=0.0$\,MHz, SIR $=0.0$\,dB'' (a non-overlapping pair)\par
Q2b: ``victim\_id: S1'' (the FM at $-$5.6\,MHz, outside the sweep)\par\smallskip
Q1: 10.6/34\quad Q2: 7.0/33\quad Q3: 5.8/33
}
\end{badbox}
\begin{goodbox}[\textcolor{white}{GPT-5.5 (ReconPilot)} \hfill Score: 65.7/100\quad|\quad 9 code calls\quad|\quad 397.1\,s]
{\scriptsize\ttfamily
\textbf{EMRB\_L5\_2005}\par\smallskip
\textbf{Q1a inventory (6 of 6):} ``\emph{I audited the ROIs with Welch PSD plus time-frequency ridge/residual inspection \ldots\ after masking its ridge, a continuous QPSK-like link remains near 7.8\,MHz}''.\par\smallskip
\textbf{Stage 3:} \texttt{consistency\_issues\_before: [``Q1b pair is not the largest non-chirp overlap in Q1a'']} $\rightarrow$ \texttt{after: []}\par\smallskip
Q1: 18.6/34\quad Q2: 22.5/33\quad Q3: 24.6/33
}
\end{goodbox}
\end{tcbraster}
\vspace{2pt}
\noindent
\begin{minipage}[t]{0.485\textwidth}
{\footnotesize\color{badframe}\textbf{Verdict:} This is a gating cascade, not weak analysis. The stored prerequisites record \texttt{q2\_victim\_identified: false} and \texttt{q3\_complete\_digital\_link\_set: false}, so Q2b, Q2c, Q2d, Q3b and Q3c collapse to zero, 43 points in total. The chirp measurement the run did base on a correctly detected source scored a full 7/7. Its own turn-3 output even printed the peak at 7.842\,MHz without ever resolving it as a source.}
\end{minipage}%
\hfill
\begin{minipage}[t]{0.485\textwidth}
{\footnotesize\color{goodframe}\textbf{Verdict:} The region map lists the same four regions the free form run found unaided, so it supplies no new detection. What it supplies is an instruction: \emph{audit the source count in every ROI, and for a broad or moving ROI inspect the residual after masking the ridge}. Following it recovered the QPSK during Stage~2, which cleared both prerequisites and turned the same downstream items into 10.5/12, 4/4, 6/8 and 7/7. The sixth source appeared only in Stage~3, when the deterministic dependency audit reported the illegal Q1b pair and the repair call measured the $-$34.99\,dBm excess that became it.}
\end{minipage}
\vspace{4pt}
\captionof{figure}{GPT-5.5 on EMRB\_L5\_2005. Under prerequisite-gated L5 scoring, a four-source inventory zeroes 43 points of otherwise competent downstream work. ReconPilot recovers the fifth source by following the region map's audit instruction and the sixth through Stage~3's dependency check, which is the division of labor the stage ablation in Section~\ref{sec:stage_ablation} predicts.}
\label{fig:case_gpt5}
\end{figure*}

\section{Benchmark Comparison}\label{app:benchmark_table}

Table~\ref{tab:benchmark_comparison} compares EMRB with representative EM/wireless and data-analysis benchmarks along three dimensions: input format, task description, and output format.
Existing EM datasets provide pre-segmented signal snippets or encoded representations and evaluate fixed classification outputs (modulation labels, device IDs, beam indices).
Data-analysis benchmarks operate on structured tables with explicit schemas.
EMRB is the only benchmark that delivers raw, full-length I/Q recordings and requires models to produce free-text engineering answers through self-directed code execution.

\begin{table*}[h]
\caption{Comparison with representative EM/wireless and data-analysis benchmarks.}
\label{tab:benchmark_comparison}
\centering\footnotesize
\setlength{\tabcolsep}{4pt}
\renewcommand{\arraystretch}{1.12}
\begin{tabularx}{\textwidth}{@{}>{\raggedright\arraybackslash}p{0.16\textwidth}>{\raggedright\arraybackslash}X>{\raggedright\arraybackslash}X>{\raggedright\arraybackslash}p{0.20\textwidth}@{}}
\toprule
\textbf{Benchmark} & \textbf{Input Format} & \textbf{Task Description} & \textbf{Output Format} \\
\midrule
\textbf{EMRB} & Raw full-length I/Q files & Open-ended EM signal analysis & Text answers \\
EM-Bench & Encoded signal-text pairs & Signal question answering & Text answers \\
RadioML & Short synthetic I/Q snippets & Modulation classification & Modulation labels \\
WiSig & WiFi I/Q packet captures & RF fingerprinting & Device IDs \\
WASD & Wireless measurements or spectrograms & Spectrum anomaly detection & Anomaly classes or regions \\
DeepSense 6G & Multimodal wireless data & Beam prediction and sensing tasks & Beams or locations \\
DA-Bench & Structured tables with questions & Tabular data analysis & Text or numeric answers \\
\bottomrule
\end{tabularx}
\end{table*}

\end{document}